\documentclass[10pt,twocolumn,letterpaper]{article}

\usepackage[pagenumbers]{cvpr} 

\usepackage{graphicx}
\usepackage{amsmath}
\usepackage{amssymb}
\usepackage{booktabs}
\usepackage{times}
\usepackage{epsfig}
\usepackage{color}
\usepackage{xcolor}
\usepackage{enumerate}
\usepackage{enumitem}
\usepackage{pifont}

\usepackage[accsupp]{axessibility}  

\usepackage{caption}
\usepackage[font=small,skip=3pt]{caption}

\usepackage[pagebackref,breaklinks,colorlinks]{hyperref}

\usepackage[capitalize]{cleveref}
\crefname{section}{Sec.}{Secs.}
\Crefname{section}{Section}{Sections}
\Crefname{table}{Table}{Tables}
\crefname{table}{Tab.}{Tabs.}

\def\cvprPaperID{8702} 
\def\confName{CVPR}
\def\confYear{2023}

\begin{document}


\title{PoseFormerV2: Exploring Frequency Domain for Efficient and Robust \\3D Human Pose Estimation}

\author{
Qitao Zhao\textsuperscript{1}$\thanks{Work was done while Qitao was an intern mentored by Chen Chen.}$ ~ 
Ce Zheng\textsuperscript{2} ~ 
Mengyuan Liu\textsuperscript{3} ~ 
Pichao Wang\textsuperscript{4} ~ 
Chen Chen\textsuperscript{2}\\ 
\small\textsuperscript{1}Shandong University \quad 
\small\textsuperscript{2}Center for Research in Computer Vision, University of Central Florida \quad
\small\textsuperscript{4}Amazon Prime Video \\
\small\textsuperscript{3}Key Laboratory of Machine Perception, Peking University, Shenzhen Graduate School \\
{\tt\small qitaozhao@mail.sdu.edu.cn} \quad {\tt\small cezheng@knights.ucf.edu} \quad {\tt\small nkliuyifang@gmail.com} \\
{\tt\small pichaowang@gmail.com} \qquad {\tt\small chen.chen@crcv.ucf.edu} 
\vspace{-10pt}
}
\maketitle

\begin{abstract}
    Recently, transformer-based methods have gained significant success in sequential 2D-to-3D lifting human pose estimation. As a pioneering work, PoseFormer captures spatial relations of human joints in each video frame and human dynamics across frames with cascaded transformer layers and has achieved impressive performance. However, in real scenarios, the performance of PoseFormer and its follow-ups is limited by two factors: (a) The length of the input joint sequence; (b) The quality of 2D joint detection. Existing methods typically apply self-attention to \textbf{all frames} of the input sequence, causing a huge computational burden when the frame number is increased to obtain advanced estimation accuracy, and they are not robust to noise naturally brought by the limited capability of 2D joint detectors. In this paper, we propose PoseFormerV2, which exploits a compact representation of lengthy skeleton sequences in the frequency domain to efficiently scale up the receptive field and boost robustness to noisy 2D joint detection. With minimum modifications to PoseFormer, the proposed method effectively fuses features both in the time domain and frequency domain, enjoying a better speed-accuracy trade-off than its precursor. Extensive experiments on two benchmark datasets (i.e., Human3.6M and MPI-INF-3DHP) demonstrate that the proposed approach significantly outperforms the original PoseFormer and other transformer-based variants. Code is released at \url{https://github.com/QitaoZhao/PoseFormerV2}.
\end{abstract}

\section{Introduction}
\label{sec:intro}

3D human pose estimation (HPE) aims at localizing human joints in 3-{dimensional} space based on monocular videos (without intermediate 2D representations) \cite{pavlakos2018ordinal, Moon_I2L_MeshNet} or 2D human joint sequences (referred to as 2D-to-3D lifting approaches) \cite{Liu_2020_CVPR, chen2020anatomy, zeng2020srnet_ECCV,wang2020motion}. With the large availability of 2D human pose detectors \cite{newell2016stacked, chen2018cascaded} plus the lightweight nature of 2D skeleton representation of humans, lifting-based methods are now dominant in 3D human pose estimation. Compared to raw monocular videos, 2D coordinates of human joints in each video frame are much more memory-friendly, making it possible for lifting-based methods to utilize a long joint sequence to boost pose estimation accuracy. 
    
\begin{figure}[t]
  \centering
   \includegraphics[width=1.0\linewidth]{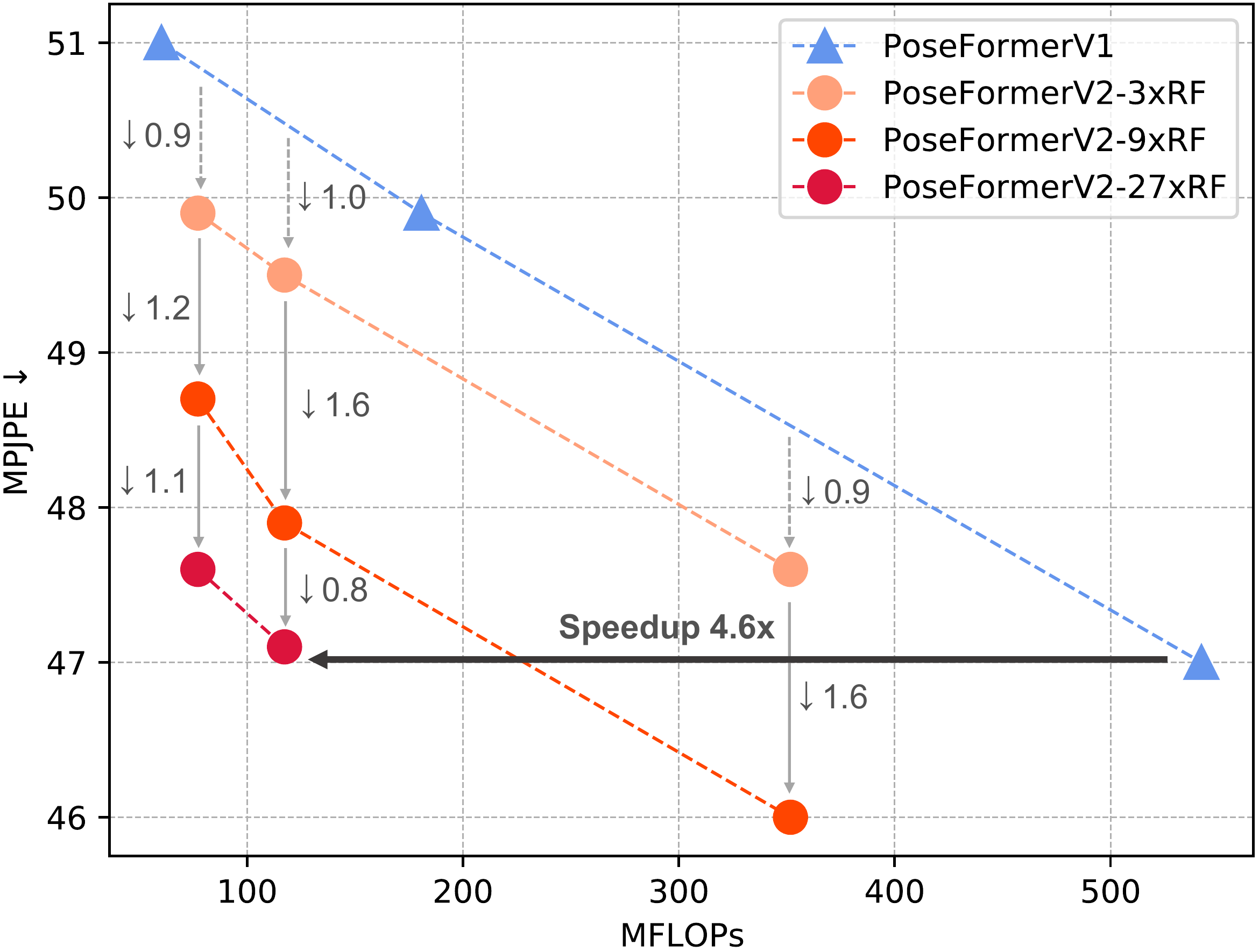}
   \caption{Comparisons of PoseFormerV2 and PoseFormerV1 \cite{Zheng_2021_ICCV} on Human3.6M \cite{Human3.6M}. RF denotes Receptive Field and $k\times$RF indicates that the ratio between the full sequence length and the number of frames as input into the spatial encoder of PoseFormerV2 is $k$, \ie, the RF of the spatial encoder is expanded by $k\times$ with a few low-frequency coefficients of the full sequence. The proposed method outperforms PoseFormerV1 by a large margin in terms of speed-accuracy trade-off, and the larger $k$ brings more significant improvements, \eg, 4.6$\times$ speedup with the $k$ of 27.}
    \vspace{-20pt}
    \label{fig:fig1}
\end{figure}
    
Transformers \cite{Attention_is_All_You_Need} first gain huge success in the field of natural language processing (NLP) \cite{devlin2018bert, brown2020language} and then extend their capacity to the computer vision community, becoming the \textit{de facto} approach for several vision tasks, \eg, image classification \cite{Dosovitskiy2020ViT, touvron2020deit, liu2021swin}, object detection \cite{carion2020end, zhu2020deformable} and video recognition \cite{Arnab_2021_ICCV, gberta_2021_ICML, yan2022multiview}. 
The discreteness of human joint representation and the requirement for long-range temporal dependency modeling in a skeleton sequence make transformers an excellent fit for lifting-based human pose estimation. Previous works \cite{Zheng_2021_ICCV, Li_2022_CVPR, zhang2022mixste, li2022exploiting,  shan2022p} have adopted transformers as the backbone for 3D human pose estimation and shown promising results.
    
\begin{figure}[t]
  \centering
   \includegraphics[width=1.0\linewidth]{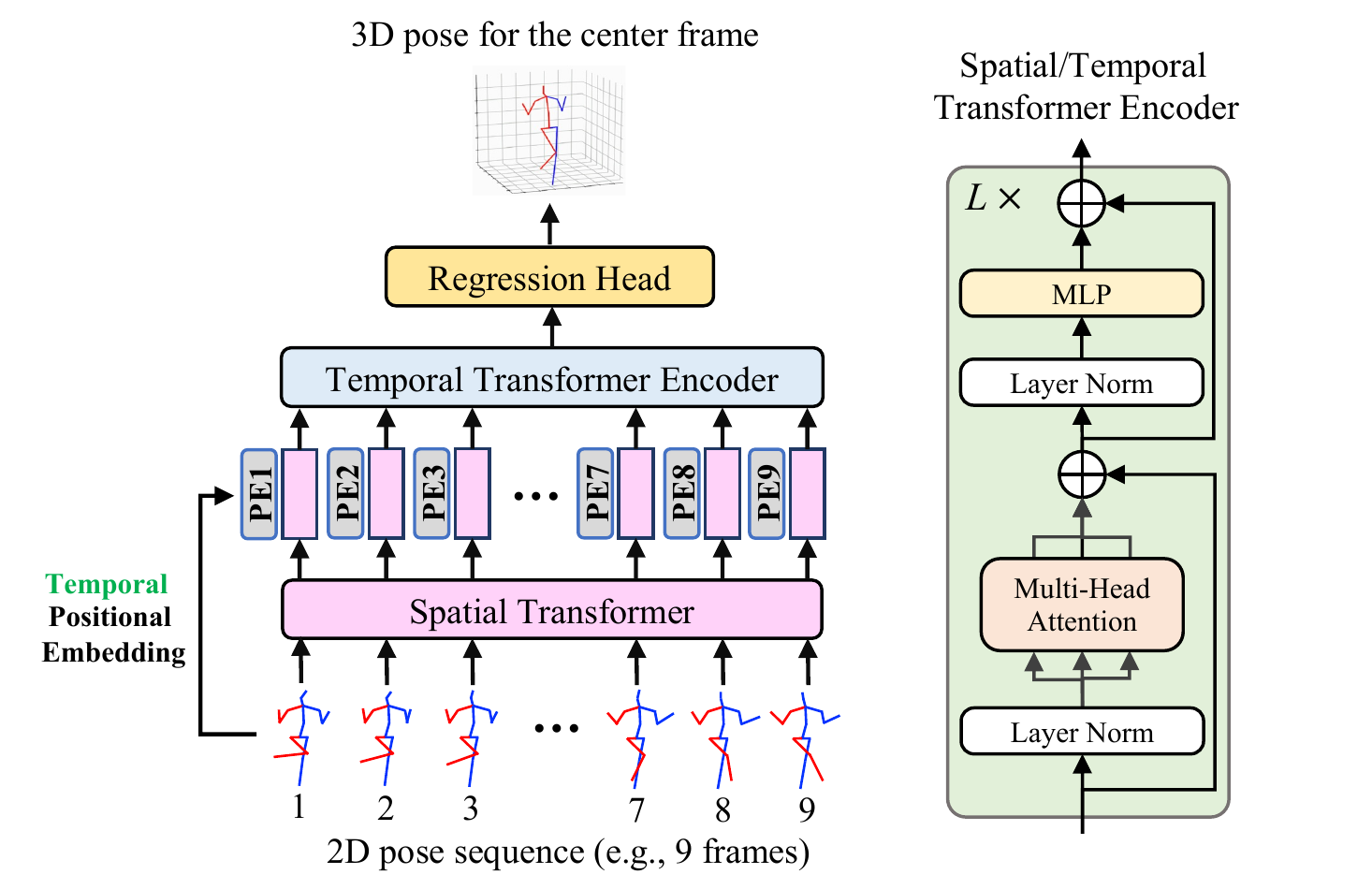}

   \caption{Overview of PoseFormerV1. PoseFormerV1 mainly consists of two modules: the spatial transformer encoder and the temporal transformer encoder. The temporal encoder of PoseFormerV1 applies self-attention to all frames given a 2D joint sequence for human motion modeling.}
   \vspace{-12pt}
   \label{fig:PoseFormerV1}
\end{figure}

As the pioneering work among transformer-based methods, PoseFormer \cite{Zheng_2021_ICCV} factorizes joint sequence feature extraction into two stages (see Fig. \ref{fig:PoseFormerV1}) and outperforms traditional convolution-based approaches. First, all joints within each frame are linearly projected into high-dimensional vectors (\ie, joint tokens) as input into the spatial transformer encoder. The spatial encoder builds up inter-joint dependencies in single frames with the self-attention mechanism. In the second stage, joint tokens of each frame are combined as one frame token, serving as input to the temporal encoder for human motion modeling across all frames in sequence. More details are included in Sec.~\ref{sec:PoseFormerV1}.
   
 \begin{table}[]
\scriptsize
\centering
\caption{The computational cost and \textbf{performance drop} brought by replacing ground-truth 2D detection with CPN \cite{chen2018cascaded} 2D pose detection for the SOTA transformer-based methods. The performance drop is reported on Human3.6M dataset (Protocol 1) \cite{Human3.6M}. RF: Receptive Field, sharing the same meaning as that in Fig. \ref{fig:fig1}.}
  {\begin{tabular}{lc|c|c|c} 
\hline
\begin{tabular}[c]{@{}c@{}} Method \end{tabular} &
\begin{tabular}[c]{@{}c@{}} \end{tabular} & 
\begin{tabular}[c]{@{}c@{}} Seq. \\Length \end{tabular} &
\begin{tabular}[c]{@{}c@{}} GFLOPs \end{tabular} & 
\begin{tabular}[c]{@{}c@{}} Perform. \\ \textbf{Drop} (mm) \end{tabular} \\ 
\hline
PoseFormerV1 \cite{Zheng_2021_ICCV} & ICCV'21 & 81 & 1.36 & 13.0   \\
StridedTransformer \cite{li2022exploiting} & TMM’22 & 243 & 1.37 & 15.2 \\
MixSTE \cite{zhang2022mixste} & CVPR'22 & 81 & 92.46 & 16.5 \\
MHFormer \cite{Li_2022_CVPR} & CVPR'22 & 81 & 3.12 & 11.8 \\
P-STMO \cite{shan2022p} & ECCV'22 & 243 & 1.74 & 13.5 \\
\hline
\textbf{PoseFormerV2} (9$\times$RF) &  & 81 & 0.35 & \textbf{8.2}  \\
\textbf{PoseFormerV2} (27$\times$RF) &  & 81 & \textbf{0.12} & 9.7  \\
\end{tabular}}
\label{tab:table1}
\vspace{-20pt}
\end{table}

Despite its capacity, the performance of PoseFormer (and other transformer-based methods) is limited by two crucial factors. \textbf{(a)} The length (number of frames) of the input 2D skeleton sequence. State-of-the-art transformer-based methods typically use extremely long sequences to obtain advanced performance, \eg, 81 frames for PoseFormer \cite{Zheng_2021_ICCV}, 243 frames for P-STMO \cite{li2022exploiting} and  351 frames for MHFormer \cite{Li_2022_CVPR}. However, densely applying self-attention to such long sequences is highly computationally expensive, \eg, the single-epoch wall-time training cost of 3-frame PoseFormer is $\sim$5 minutes while for 81-frame PoseFormer the cost surges to $\sim$1.5 hour on an RTX 3090 GPU. \textbf{(b)} The quality of 2D joint detection. 2D joint detectors inevitably introduce noise due to bias in their training dataset and the temporal inconsistency brought by the single-frame estimation paradigm. For example, PoseFormer achieves 31.3mm MPJPE (Mean Per Joint Position Error) using the \textbf{ground-truth} 2D detection on the Human3.6M dataset \cite{Human3.6M}. This result drops significantly to 44.3mm when the clean input is replaced by the CPN \cite{chen2018cascaded} 2D pose detection. In practice, the long-sequence inference may be unaffordable for hardware deployment on resource-limited devices such as AR/VR headsets and high-quality 2D detection is hard to obtain. More quantitative results about the efficiency to process long sequences and the robustness to noisy 2D joint detection of existing transformer-based methods are available in Table \ref{tab:table1}.
    
Driven by these practical concerns, we raise two important research questions: 
\vspace{-5pt}
\begin{itemize}[noitemsep,leftmargin=*]  
\item \textit{Q1: How to efficiently utilize long joint sequences for better estimation precision?}
\item \textit{Q2: How to improve the robustness of the model against unreliable 2D pose detection?}
\end{itemize}
\vspace{-5pt}
Few works have tried to answer either of these two questions by incorporating hand-crafted modules, \eg, the downsampling-and-uplifting module \cite{einfalt_up3dhpe_WACV23} that only processes a proportion of video frames for improved efficiency, the multi-hypothesis module \cite{Li_2022_CVPR} to model the depth ambiguity of body parts and the uncertainty of 2D detectors. \emph{However, none of them manages to find a single solution to these two questions simultaneously, and even worse, a paradox seemingly exists between solutions to the questions above, \eg, multiple hypotheses} \cite{Li_2022_CVPR} \emph{improve robustness but bring additional computation cost (see also Table \ref{tab:table1}).}

In this paper, we present our initial attempt to \emph{``kill" two birds with one stone}. With restrained modifications to the prior art PoseFormer, we show that the appropriate form of representation for input sequences might be the key to answering these questions simultaneously. Specifically, we shed light on the barely explored frequency domain in 3D HPE literature and propose to encode the input skeleton sequences into low-frequency coefficients. The insight behind this representation is surprisingly simple: On the one hand, low-frequency components are enough to represent the entire visual identity \cite{Xu_2020_CVPR, wang2022vtclfc} (\eg, 2D images in image compression and joint trajectories in this case), thus removing the need for expensive all-frame self-attention; On the other, the low-frequency representation of the skeleton sequence itself filters out high-frequency noise (jitters and outliers) \cite{mao2019learning, mao2020history} contained in detected joint trajectories. 

We inherit the spatial-temporal architecture from PoseFormer but force the spatial transformer encoder to only ``see'' a few central frames in a long sequence. Then we complement ``short-sighted'' frame-level features (the output of the spatial encoder) with global features from low-frequency components of the complete sequence. Without resorting to the expensive frame-to-frame self-attention for all time steps, the temporal transformer encoder is reformulated as a Time-Frequency Feature Fusion module.


Extensive experiments on two 3D human pose estimation benchmarks (\ie, Human3.6M \cite{Human3.6M} and MPI-INF-3DHP \cite{MPIINF}) demonstrate that the proposed approach, dubbed as \textbf{PoseFormerV2}, significantly outperforms its precursor (see Fig. \ref{fig:fig1}) and other transformer-based variants in terms of speed-accuracy trade-off and robustness to noise in 2D joint detection. Our \textbf{contributions} are three-fold:
    
\setlist{nolistsep}
\begin{itemize}[noitemsep,leftmargin=*]  
\item To the best of our knowledge, we are the first to utilize a frequency-domain representation of input joint sequences for 2D-to-3D lifting HPE. We find this representation an ideal fit to concurrently solve two important issues in the field (\ie, the efficiency to process long sequences and the robustness to unreliable joint detection), and experimental evidence shows that this approach can easily generalize to other models.

\item We design an effective Time-Frequency Feature Fusion module to narrow the gap between features in the time domain and frequency domain, enabling us to strike a flexible balance between speed and accuracy.

\item Our PoseFormerV2 outperforms other transformer-based methods in terms of the speed-accuracy trade-off and robustness on Human3.6M and achieves the state-of-the-art on MPI-INF-3DHP. 
   
\end{itemize}

\section{Related Work}
\label{sec:related works}

Our method is built on conceptually simple PoseFormer \cite{Zheng_2021_ICCV}, and we aim at improving its efficiency to operate long sequences and its robustness to noisy joint detection from a frequency-domain perspective. Therefore, here we mainly focus on this line of works (transformer-based methods) in 2D-to-3D lifting HPE and introduce applications of frequency domain representations in computer vision literature, especially in skeleton-based tasks that are most related to lifting-based 3D HPE.

\subsection{Transformer-based 3D Human Pose Estimation}
PoseFormer \cite{Zheng_2021_ICCV} is the first work to adopt the vision transformer as the backbone network in lifting-based 3D human pose estimation, and it outperforms previous CNN-based methods by a large margin. Though being competitive, Zhang \etal \cite{zhang2022mixste} point out that the spatial-then-temporal paradigm of PoseFormer may neglect distinct temporal patterns for each joint, and propose to adopt alternate spatial-temporal transformer layers for fine-grained joint-specific feature extraction. MHFormer \cite{Li_2022_CVPR} further incorporates task-related prior knowledge into transformers for 3D HPE. Specifically, 2D-to-3D lifting is an inverse problem where more than one reasonable solutions exist, therefore they generate multiple hypotheses to model ambiguous body parts and uncertainty in joint detectors, achieving advanced performance. Inspired by the progress of Masked Image Modeling (MIM) in image classification \cite{he2022masked, wei2022masked, xie2022simmim}, P-STMO \cite{shan2022p} applies Masked Joint Modeling to 3D HPE with self-supervised learning.

Another line of works \cite{li2022exploiting, einfalt_up3dhpe_WACV23} improves the efficiency of transformer-based methods. Taking advantage of the temporal redundancy in 2D joint sequences, StridedTransformer \cite{li2022exploiting} replaces the parameter-heavy fully-connected layers with strided convolutions. Einfalt \etal \cite{einfalt_up3dhpe_WACV23} claim that the per-frame 2D joint detection is even more computationally expensive than lifting models themselves and propose to downsample input video frames with a fixed interval and adopt the 2D joint detector and lifting model only on these sampled frames. While being more efficient than previous works, aforementioned methods \cite{li2022exploiting, einfalt_up3dhpe_WACV23} reduce participants in self-attention along the temporal dimension utilizing only the consistency in adjacent video frames rather than from a global view, and therefore they may suffer from a considerable performance drop.

\subsection{Frequency Representation in Vision}
Since the human visual system is more sensitive to low-frequency components of images, traditional image compression algorithms, \eg, JPEG \cite{pennebaker1992jpeg} and JPEG 2000 \cite{skodras2001jpeg}, reduce memory cost to store 2D images by allocating more storage budget to low-frequency Discrete Cosine Transform (DCT) coefficients of the image. Following the same logic, \cite{Xu_2020_CVPR} proposes to adaptively remove uninformative channels of DCT components for 2D images to boost image classification efficiency. More recently, some works \cite{rao2021global, guibas2021adaptive} propose to replace the costly self-attention mechanism with frequency transforms that can be accelerated by their fast algorithms. GFN \cite{rao2021global} proposes to efficiently mix visual tokens with learnable frequency filters, and AFNO \cite{guibas2021adaptive} further improves the performance of token mixer in the frequency domain with operator learning. Moreover, Wang \etal \cite{wang2022vtclfc} utilize low-frequency Fast Fourier Transform (FFT) components to compress vision transformers.

\textbf{Skeleton-based tasks} are more relevant to our work that takes 2D skeleton sequences as input. In the human motion prediction literature, previous works \cite{mao2019learning, mao2020history} transform the skeleton sequence from the time domain into DCT coefficients to encode human dynamics as compared to static joint coordinates. They observe that discarding a few high-frequency coefficients does not necessarily bring a performance drop but even improves the smoothness of predicted future motions. However, frequency-domain representations of 2D joint sequences have not yet been explored in lifting-based 3D human pose estimation.

Our approach is inspired by these former attempts of applying frequency transforms to vision tasks but from a different view. We include more details about our motivations to choose the DCT coefficient representation in Sec.~\ref{sec: freq}.

\section{Method}
\label{sec:method}

PoseFormer \cite{Zheng_2021_ICCV} facilitates 3D human pose estimation by factoring sequence feature extraction into two stages, \ie, the spatial encoder and temporal encoder, which has proved to be effective. However, it suffers from a huge computational burden when the length of input sequences is increased, and is sensitive to noisy joint detection. In this section, we introduce the details of PoseFormerV2 which utilizes a frequency representation of the input sequence to overcome two aforementioned problems.

\subsection{Preliminaries of PoseFormerV1 \cite{Zheng_2021_ICCV}}
\label{sec:PoseFormerV1}

We start by giving a brief overview of PoseFormerV1 (see Fig. \ref{fig:PoseFormerV1}), laying the basis for the discussions on its improvements. PoseFormerV1 consists of two main modules, the spatial encoder for single-frame joint correlation modeling and the temporal encoder for cross-frame human motion modeling. Given an input 2D skeleton sequence $\mathbf{x} \in \mathbb{R} ^{F \times J \times 2}$, where ${F}$ denotes the sequence length and ${J}$ denotes the joint number of the human representation. First, coordinates of all joints of a person in each frame are linearly projected to a ${c}$-{dimensional} vector (\ie, joint embedding) denoted as $\mathbf{z}_0 \in \mathbb{R} ^{F \times J \times c}$. A learnable spatial positional embedding $\mathbf{E}_{SPos} \in \mathbb{R} ^{1 \times J \times c}$ \cite{Dosovitskiy2020ViT} is added to $\mathbf{z}_0$ to encode joint-dependent information.

\textbf{Spatial transformer encoder} builds up spatial dependencies for joint embeddings of each frame $\mathbf{z}_{0}^i \in \mathbb{R} ^{1 \times J \times c}$ individually with the self-attention mechanism. In this stage, the number of tokens fed into each transformer block is ${J}$. The output of the spatial transformer encoder of $L$ layers for the $i$-{th} frame is denoted by $\mathbf{z}_{L}^i \in \mathbb{R} ^{1 \times J \times c}$. Then per-frame representations are flattened and concatenated as input $\mathbf{Z}_0 \in \mathbb{R} ^{F \times (J \cdot c)}$ into the temporal transformer encoder.

\textbf{Temporal transformer encoder}. Similarly, the input $\mathbf{Z}_0$ is added with a learnable temporal positional embedding $\mathbf{E}_{TPos} \in \mathbb{R} ^{F \times (J \cdot c)}$ to encode index-dependent information for each frame. The temporal encoder with $M$ transformer layers densely models frame-to-frame dependencies across the whole sequence, and its output is denoted by $\mathbf{Z}_M \in \mathbb{R} ^{F \times (J \cdot c)}$. In this stage, the token number for each transformer layer is ${F}$, which is the input sequence length.

\textbf{Regression head}. To estimate the 3D pose of the central frame in sequence, a simple 1D convolution is used to gather temporal information and a linear projection outputs the final pose representation $\mathbf{y} \in \mathbb{R} ^{1 \times (J\cdot 3)}$.

\textbf{Limitations of PoseFormerV1.} Modeling joint dependencies within each frame and human motions across frames with transformer layers is straightforward. While such dense modeling brings advanced estimation accuracy, it is computationally unfriendly due to the quadratic computation growth of self-attention with respect to the token number (\ie, the joint number in the spatial encoder and the sequence length in the temporal encoder) especially when the input sequence length is increased. Although the token number for spatial transformer layers (\ie, the joint number) is independent of the frame number, it is worth noting that the sequence length implicitly affects the computational budgets of the spatial encoder in real scenarios because of the limited parallelization ability of GPUs. In addition to the efficiency issue, PoseFormerV1 is sensitive to the quality of input 2D joint detection (experimental evidence is available in Table \ref{tab:table1} and Sec.~\ref{sec:comparison}).


In the following, we present an alternative solution to overcome the limitations of PoseFormerV1 with the frequency-domain representation of the input sequence. 

\begin{figure}[t]
  \centering
   \includegraphics[width=1.0\linewidth]{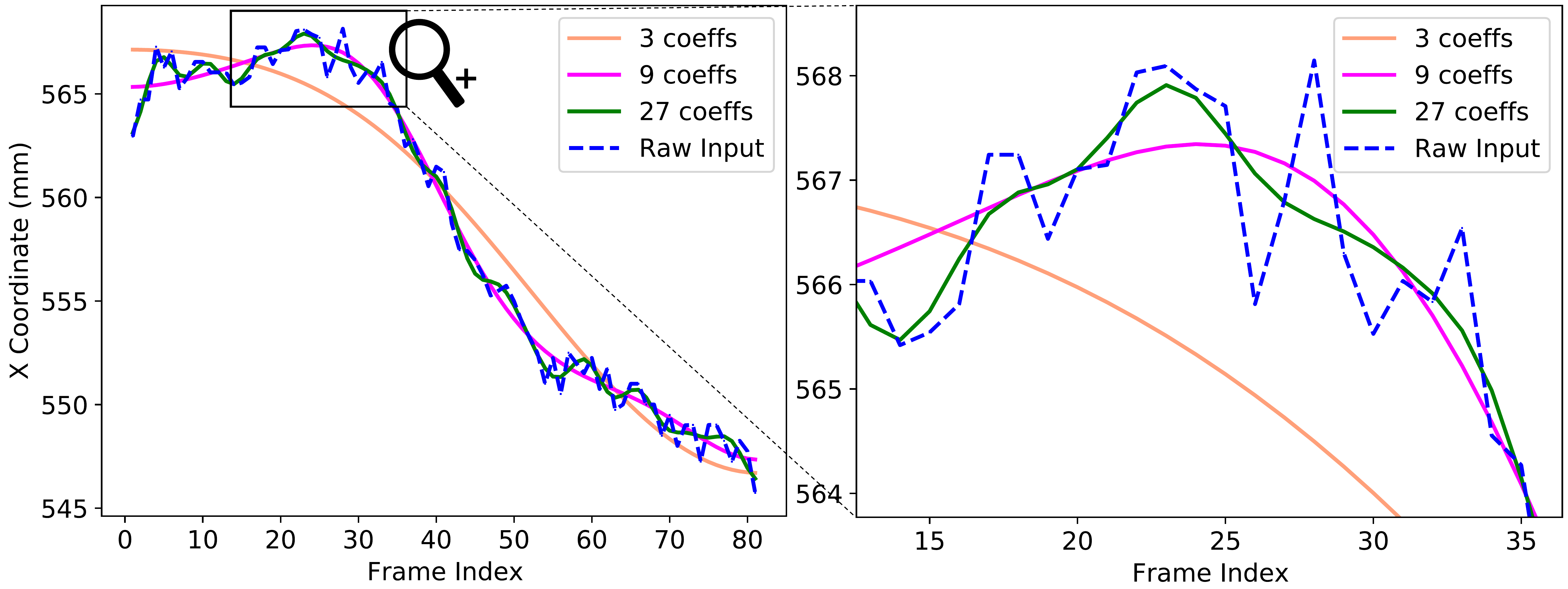}
   \caption{A randomly selected example of the CPN-detected \cite{chen2018cascaded} joint trajectory in Human3.6M \cite{Human3.6M} and its reconstructions with first 3, 9, and 27 DCT coefficients (81 in total). Note that even with only the first 3 coefficients, the reconstructed (orange) curve captures the overall characteristics of the raw input, and is smoother.}
   \vspace{-15pt}
   \label{fig:DCT}
\end{figure}
\begin{figure*}[t]
  \centering
   \includegraphics[width=0.95\linewidth]{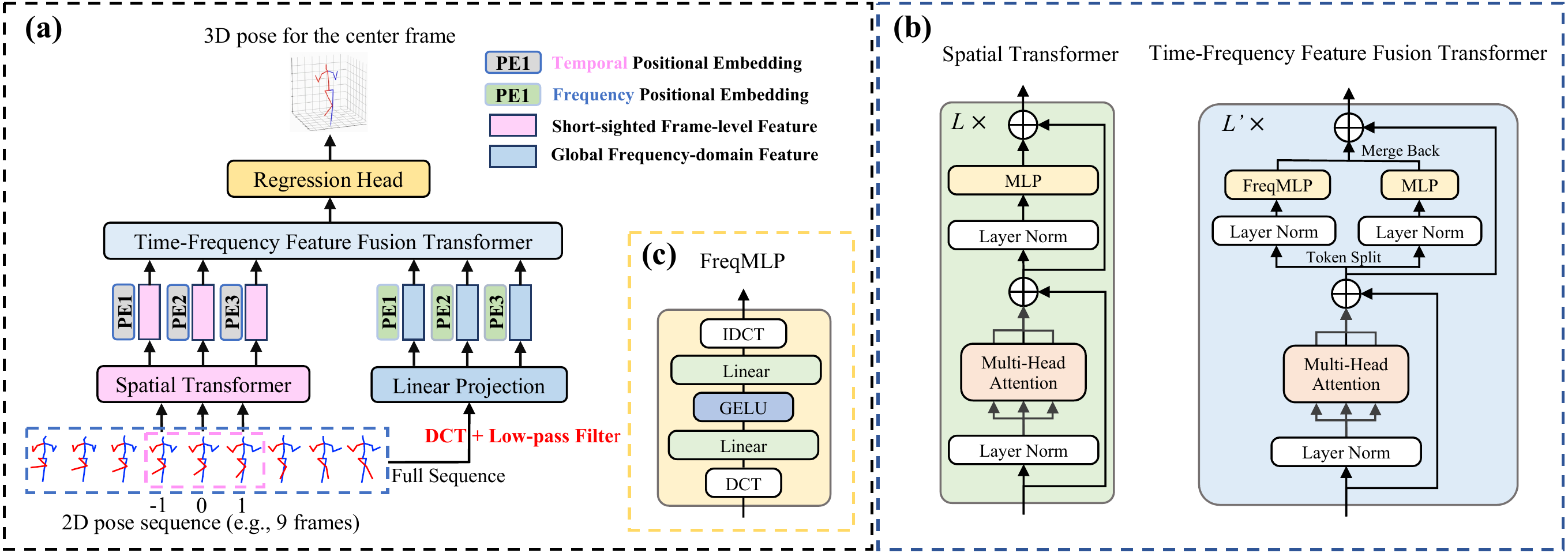}

   \caption{(a) Overview of \textbf{PoseFormerV2}. (b) Spatial Transformer and Time-Frequency Feature Fusion Transformer. (c) FreqMLP (Frequency Multi-Layer Perceptron). To exemplify, in (a) we use 3 central frames (index -1, 0, and 1) for fine-grained frame-level feature extraction and the first 3 DCT coefficients of the full 9-frame sequence for global frequency-domain feature extraction. Therefore, the effective number of frames as input to the spatial encoder and temporal encoder is reduced compared to PoseFormerV1 (3,6 \vs 9,9).}
   \vspace{-15pt}
   \label{fig:framework}
\end{figure*}
\subsection{PoseFormerV2}
\subsubsection{Frequency Representation of Skeleton Sequence}
\label{sec: freq}

\textbf{Motivation.} We propose to transform the input skeleton sequence into the frequency domain with Discrete Cosine Transform (DCT) and utilize only a portion of low-frequency coefficients. DCT coefficients encode multiple levels of temporal information for the input time series. Specifically, low-frequency coefficients encode its rough contour while high-frequency ones encode its details, \eg, jitters or sharp changes. To better illustrate our motivation to choose this representation, we provide an 81-frame example of the CPN-detected \cite{chen2018cascaded} joint trajectory of action ``Directions'' in the Human3.6M \cite{Human3.6M} dataset and its reconstructions with first 3, 9, and 27 DCT coefficients respectively (see Fig. \ref{fig:DCT}). As the number of kept DCT coefficients increases, the reconstructed trajectory becomes closer to the raw input but less smooth. Note that with only 3 DCT coefficients (denoted by the orange curve), the overall trend of the original trajectory is captured, and with 9 and 27 coefficients (pink and green curves), the characteristics of the raw sequence are better preserved while high-frequency noise (zig-zags) is removed. These observations motivate us to exploit a few highly informative low-frequency DCT components of the input joint sequence as the \textit{compact and denoised sequence representation} in our work. With such representation, we significantly reduce the effective length of the sequence as input and promote the robustness of our model against the noise contained in 2D joint detection. We include a formal introduction to DCT in \textcolor{blue}{supplementary}.

\subsubsection{Architecture}
\label{sec:PoseFormerV2}

In this part, we introduce the architecture of the proposed approach, PoseFormerV2 (see Fig. \ref{fig:framework} for an overview). 

\textbf{Spatial transformer encoder.} Given a 2D skeleton sequence $\mathbf{x} \in \mathbb{R} ^{F \times J \times 2}$ (preferably a long sequence, \eg, $F$ is 81), we first sample $F'$ (typically $F' \ll F$) frames around the sequence center (the frame of index 0 in Fig. \ref{fig:framework} (a)), denoted by $\mathbf{x'} \in \mathbb{R} ^{F' \times J \times 2}$, as input to the spatial encoder. The output of the spatial encoder is denoted by $\mathbf{z}^{Time} \in \mathbb{R} ^{F' \times (J \cdot c)}$ (frame-level features in the time domain). The design of the spatial encoder directly follows PoseFormerV1.

\textbf{Low-frequency DCT coefficients.} $\mathbf{z}^{Time}$ is referred to as ``short-sighted'' because its receptive field ($F'$) is restricted in comparison to the entire sequence length ($F$). To efficiently exploit the long-range human dynamics of the original sequence, we resort to its frequency-domain representation. We first convert the full sequence $\mathbf{x} \in \mathbb{R} ^{F \times J \times 2}$ into DCT coefficients, denoted by $\mathbf{C} \in \mathbb{R} ^{F \times J \times 2}$. Then we keep only the first $N$ ($\ll F$) coefficients $\mathbf{C'} \in \mathbb{R} ^{N \times J \times 2}$ using a low-pass filter for every joint trajectory,  where temporal information of the original sequence is largely maintained and high-frequency noise is removed. Low-frequency coefficients $\mathbf{C'}$ are flattened and linearly projected to $\mathbf{z}^{Freq} \in \mathbb{R} ^{N \times (J \cdot c)}$ (the embedding of frequency coefficients). $\mathbf{z}^{Freq}$ is summed with a learnable frequency positional embedding $\mathbf{E}_{FPos}$ (like $\mathbf{E}_{TPos}$ in PoseFormerV1). Features from both the time domain and frequency are concatenated together, formulated as
\vspace{-2pt}
\begin{equation}
    \mathbf{z} = [\mathbf{z}^{Time}; \mathbf{z}^{Freq}],
    \vspace{0pt}
\end{equation}
fed to the Time-Frequency Feature Fusion module. 

\textbf{Time-Frequency Feature Fusion.}
We adopt transformer layers for cross-frame temporal dependency modeling as in PoseFormerV1. Compared to PoseFormerV1 which entirely extracts features in the time domain, the proposed method fuses features from both the time domain and frequency domain. To narrow the gap between the two domains, we introduce simple modifications to vanilla transformer layers. \textbf{(1)} Time-domain and frequency-domain features share self-attention but use separate feed-forward networks; \textbf{(2)} We apply FreqMLP (Frequency Multi-Layer Perceptron) in the feed-forward networks for time-domain features $\mathbf{z}^{Time}$ (see Fig. \ref{fig:framework} (b)(c)). In our FreqMLP, we utilize DCT and IDCT before and after the vanilla MLP. The intuition behind this approach is: High-frequency noise is filtered out of frequency domain features with a low-pass filter, but detailed human motion features (\eg, fast local motions) may also be lost as noise. To address this issue, FreqMLP acts as a trainable frequency-domain filter, allowing us to adaptively adjust the weight of each frequency component in the embedding of 2D joint coordinates (\ie, time-domain features), being a complement to frequency features. These modules are formulated as: 
\vspace{-5pt}
\begin{align}
\small
    &\mathbf{z}^{'}_{k} = {\rm MSA}(\mathbf{z}_{k}), \\
    &\mathbf{z}_{k}^{Time}, \mathbf{z}_{k}^{Freq} = \mathbf{z}^{'}_{k}[:F'], \mathbf{z}^{'}_{k}[F':], \\
    &\mathbf{z}_{k+1} = {\rm Concat}({\rm FreqMLP}(\mathbf{z}_{k}^{Time}), {\rm MLP}(\mathbf{z}_{k}^{Freq})), 
    \vspace{-5pt}
\end{align}
where \rm MSA denotes Multi-head Self-Attention and $F'$ is the number of sampled central frames. The effectiveness of the aforementioned modifications is verified in Sec.~\ref{sec:ablation}. It's important to recognize that the concatenation operation results in a higher number of tokens for the transformer. However, by restricting the spatial encoder to only observe a limited number of central frames and incorporating a small percentage of low-frequency DCT coefficients to expand its receptive field, we can decrease the overall computation in a flexible manner. This approach not only reduces computational costs but also enhances the model's resistance to noise compared to PoseFormerV1.

\textbf{Regression head and loss function.} Following PoseFormerV1, we use the 1D convolution layer to gather temporal information and a linear projection to obtain the final 3D pose $\mathbf{y} \in \mathbb{R} ^{1 \times (J\cdot 3)}$ for the central frame of the sequence. We use the standard MPJPE (Mean Per Joint Position Error) loss as PoseFormerV1 to train our model.

\section{Experiments}
\label{sec:experiments}

\subsection{Datasets and Evaluation Metrics}
We conduct experiments on two 3D human pose estimation datasets, \ie, Human3.6M \cite{Human3.6M} and MPI-INF-3DHP \cite{MPIINF} to demonstrate the effectiveness of our method. More detailed descriptions of both datasets and their respective evaluation metrics are in the \textcolor{blue}{supplementary}.

\subsection{Implementation Details and Analysis}
\label{sec:ID}
The proposed method includes three important hyper-parameters that are specific to experimental settings. These include the number of frames ($f$) used as input in the spatial encoder, the length of the entire input sequence ($F$) representing the enlarged receptive field, and the number of kept DCT coefficients ($n$) utilized to incorporate long-range temporal information. If not specified, we simply set $n$ = $f$ for convenience. In practice, they can be further tuned for a flexible speed-accuracy trade-off. When $f$ equals 1, $n$ is set to 3 because a single DCT coefficient may be insufficient to encode temporal information from lengthy input sequences. As $f$ and $n$ are fixed, the computational complexity of the model is predetermined (\ie, the token number for the spatial encoder and that for the feature-fusion module are fixed). We may vary $F$ to effectively expand the model's receptive field from a limited $f$ to an arbitrary value, bringing no additional computational overhead. This enables us to efficiently use long sequences to improve accuracy. We provide details of the hyper-parameters for model architecture and training in the \textcolor{blue}{supplementary}.
 
\begin{table}[]
\scriptsize
\centering
\caption{Quantitative comparisons with previous transformer-based methods on Human3.6M (in mm). $f$: number of frames as input to the model, Seq. Len.: length of the entire input sequence (\ie, the effective Receptive Field). The best scores are marked in bold. (*) indicates using an additional pre-training stage and (\dag) indicates our re-implementation.}
\vspace{0pt}
{
\begin{tabular}{l|c|c|c|l}
\begin{tabular}[l]{@{}c@{}} Method \end{tabular} &
\begin{tabular}[c]{@{}c@{}} $f$ \end{tabular} & 
\begin{tabular}[c]{@{}c@{}} Seq. \\Len. \end{tabular} &
\begin{tabular}[c]{@{}c@{}} MFLOPs \end{tabular} & 
\begin{tabular}[c]{@{}c@{}} MPJPE $\downarrow$ / \\P-MPJPE $\downarrow$ \end{tabular} \\
\hline
PoseFormerV1 \cite{Zheng_2021_ICCV} ICCV'21 & 27 & 27 & 542.1 & 47.0/--  \\
StridedTrans. \cite{li2022exploiting} TMM'22 & 81 & 81 & 342.5 & 47.5/--  \\
MixSTE \cite{zhang2022mixste}(\dag) CVPR'22 & 3 & 3 & 3420 & 49.6/38.9  \\
MHFormer \cite{Li_2022_CVPR} CVPR'22 & 9 & 9 & 342.9 & 47.8/--  \\
MHFormer \cite{Li_2022_CVPR} CVPR'22 & 27 & 27 & 1031.8 & 45.9/--  \\
P-STMO \cite{shan2022p}(*) ECCV'22 & 27 & 81 & 163 & 46.8/--  \\
P-STMO \cite{shan2022p}(*) ECCV'22 & 81 & 81 & 493 & 45.6/--  \\
Einfalt \etal \cite{einfalt_up3dhpe_WACV23} WACV'23 & 9 & 81 & 543 & 47.9/-- \\ \hline
\textbf{PoseFormerV2} & 1 & 9 & \textbf{77.2}  & 49.9/38.7  \\ 
\textbf{PoseFormerV2} & 1 & 27 & \textbf{77.2} & 48.7/37.8  \\ 
\textbf{PoseFormerV2} & 1 & 81 & \textbf{77.2} & \textbf{47.6}/\textbf{37.3}  \\ \hline
\textbf{PoseFormerV2} & 3 & 9 & 117.3 & 49.5/38.5  \\ 
\textbf{PoseFormerV2} & 3 & 27 & 117.3 & 47.9/37.4  \\ 
\textbf{PoseFormerV2} & 3 & 81 & 117.3 & \textbf{47.1}/\textbf{37.3}  \\ \hline
\textbf{PoseFormerV2} & 9 & 27 & 351.7 & 47.6/37.1  \\ 
\textbf{PoseFormerV2} & 9 & 81 & 351.7 & \textbf{46.0}/\textbf{36.1}  \\ \hline
\textbf{PoseFormerV2} & 27 & 243 & 1054.8 & \textbf{45.2}/\textbf{35.6}  \\ 
\end{tabular}
}
\vspace{-10pt}
\label{tab:Human3.6M}
\end{table}
\begin{figure}[t]
  \centering
   \includegraphics[width=0.96\linewidth]{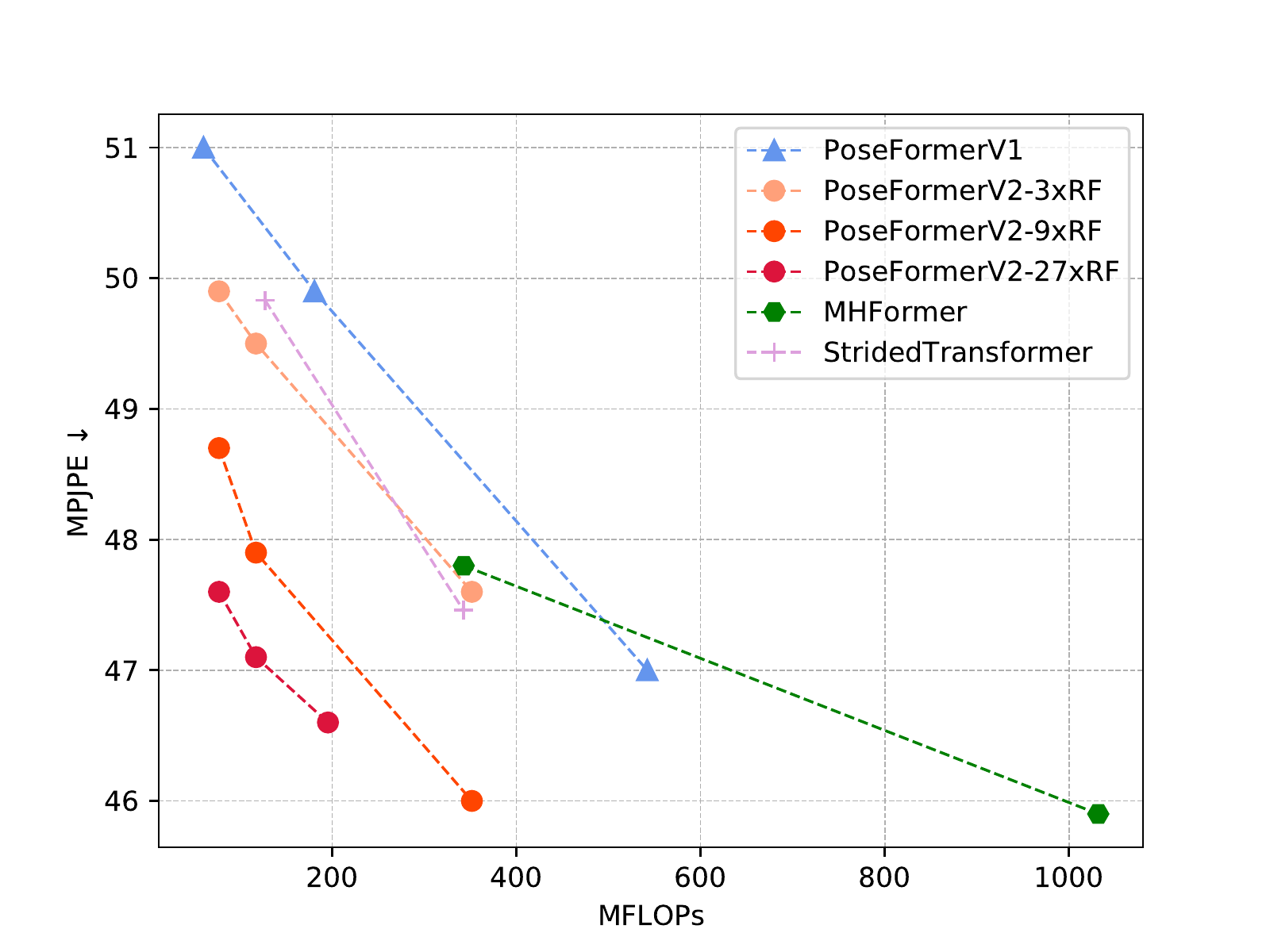}
   \vspace{-0pt}
   \caption{Comparisons of PoseFormerV2 and other state-of-the-art transformer-based methods on Human3.6M (in mm). RF: Receptive Field and $k\times$RF indicates that the RF of PoseFormerV2 is expanded by $k\times$ with a few low-frequency DCT coefficients of the full sequence. The proposed approach outperforms others in terms of speed-accuracy trade-off, and the larger $k$, the larger improvements over other methods. (Best viewed in color)}
   \label{fig:Human3.6M}
   \vspace{-10pt}
\end{figure}
\begin{figure}[t]
\vspace{-2pt}
  \centering
   \includegraphics[width=0.96\linewidth]{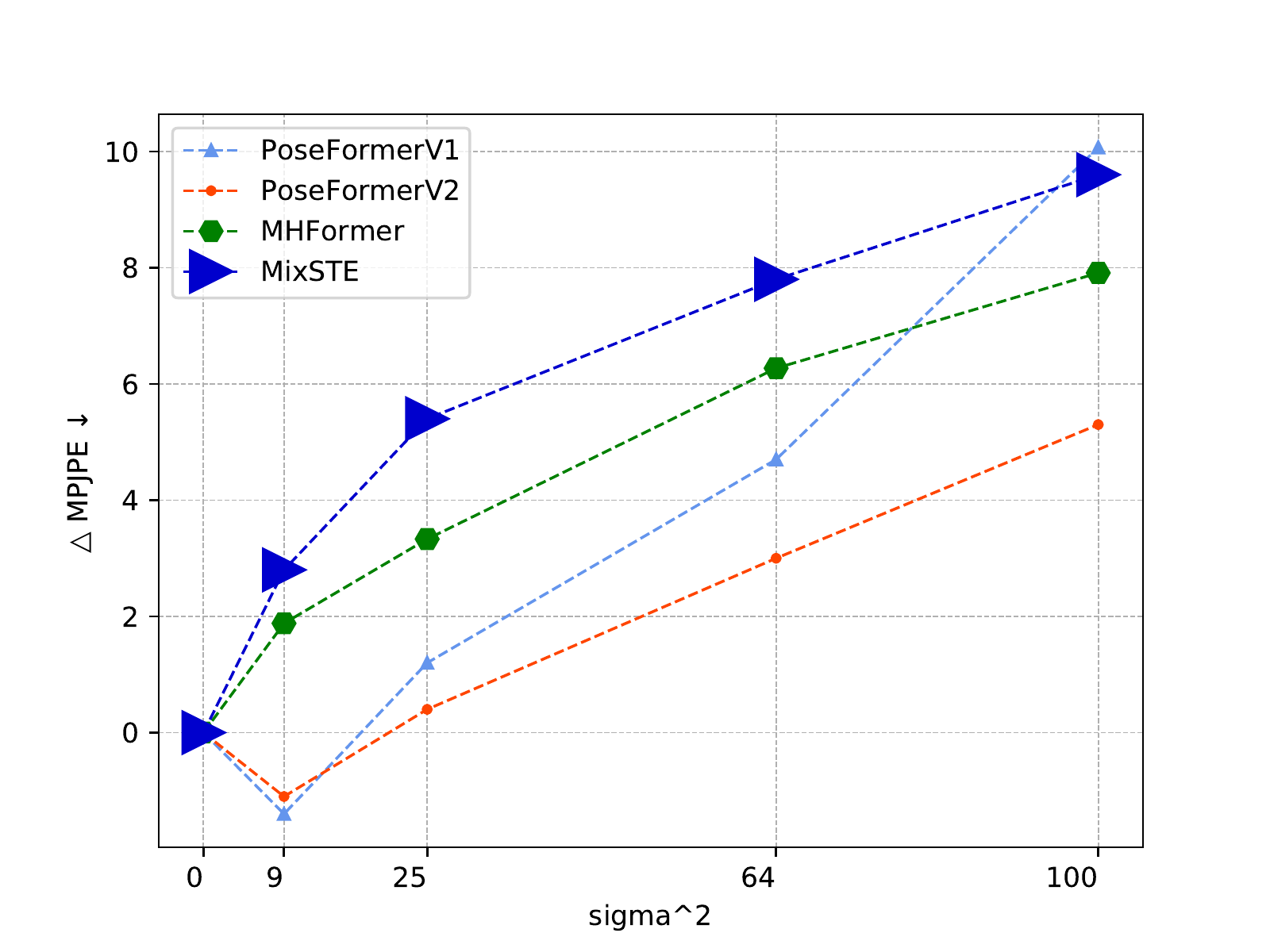}
   \caption{Comparisons of PoseFormerV2 and other transformer-based methods \cite{Zheng_2021_ICCV, Li_2022_CVPR, zhang2022mixste} in terms of robustness to noise on Human3.6M. Zero-mean Gaussian noise of standard deviation $sigma$ is added to ground truth 2D detection, and we show their performance drop ($\triangle $MPJPE, in mm) as $sigma$ increases. \textcolor{blue}{The size of markers indicates the computational cost of models.}}
   \vspace{-5pt}
   \label{fig:noise}
\end{figure}
\begin{table}[]
\scriptsize
\centering
\caption{Quantitative comparisons with previous methods on MPI-INF-3DHP. $T$: the entire sequence length, 1 by default. The best scores are marked in bold. (*) indicates using an additional pre-training stage and (\dag) indicates our re-implementation.}
\vspace{-0pt}
{
\begin{tabular}{l|c|ccc}

Method & {} & PCK $\uparrow$ & AUC $\uparrow$ & MPJPE $\downarrow$  \\ \hline
Mehta \etal \cite{MPIINF} & 3DV'17                 & 75.7 & 39.3 & 117.6 \\
Mehta \etal \cite{mehta2017vnect} & ACM ToG'17             & 76.6 & 40.4 & 124.7 \\
Pavllo \etal \cite{pavllo2019} ($T$=81)   & CVPR'19  & 86.0 & 51.9 & 84.0  \\
Pavllo \etal \cite{pavllo2019} ($T$=243)   & CVPR'19 & 85.5 & 51.5 & 84.8  \\
Lin \etal \cite{lin2019trajectory} ($T$=25) & BMVC'19                  & 83.6 & 51.4 & 79.8  \\ 
Li \etal \cite{Li_2020_CVPR} & CVPR'20   
               & 81.2 & 46.1 & 99.7  \\
Chen \etal \cite{chen2020anatomy} ($T$=81)  & TCSVT'21   & 87.9 & 54.0 & 78.8  \\ 
PoseFormerV1 \cite{Zheng_2021_ICCV} ($T$=9)(\dag) & ICCV'21 &  95.4  & 63.2 & 57.7  \\ 
MHFormer \cite{Li_2022_CVPR} ($T$=9) & CVPR'22 &  93.8  & 63.3 & 58.0  \\  
MixSTE \cite{zhang2022mixste} ($T$=27) & CVPR'22 &  94.4  & 66.5 & 54.9  \\  
P-STMO \cite{shan2022p} ($T$=81)(*) & ECCV'22 & \textbf{97.9} & 75.8 & 32.2  \\     \hline
\textbf{PoseFormerV2} ($T$=81) &     & \textbf{97.9} & \textbf{78.8} & \textbf{27.8}  \\
\end{tabular}
}
\vspace{-5pt}
\label{tab:mpi}
\end{table}

 \subsection{Comparisons with State-of-the-art Methods}
 \label{sec:comparison}
 
 \textbf{Human3.6M.} We compare our method with PoseFormerV1 and other transformer-based methods on Human3.6M (Table \ref{tab:Human3.6M}). We demonstrate the flexibility of our model by varying the value of $f$ and the sequence length. Our method is particularly efficient when the expanding ratio (\ie, the ratio of full sequence length to $f$) is large. For example, with an expanding ratio of 81, it achieves 47.6mm MPJPE with only 77.2 MFLOPs as compared to the 47.8mm MPJPE of MHFormer \cite{Li_2022_CVPR} with 342.9 MFLOPs ({4.4$\times$ slower}). Moreover, with a similar computational budget (around 350 MFLOPs) and the same full sequence length (81), our method achieves 46.0mm MPJPE whereas StridedTransformer \cite{li2022exploiting} obtains 47.5mm MPJPE (3.2\%$\uparrow$). Fig.~\ref{fig:Human3.6M} presents a clearer comparison, showing that the proposed method outperforms other transformer-based methods in terms of speed-accuracy trade-off. Note that the methods with an additional pre-training stage and computationally heavy MixSTE \cite{zhang2022mixste} (3420 MFLOPs for only 3-frame input) are not included. The improvements of PoseFormerV2 over PoseFormerV1 are provided in Fig.~\ref{fig:fig1}.
 
 In order to demonstrate that the inclusion of low-frequency DCT coefficients helps improve the robustness of the proposed method, we make the lifting-based pose estimation task more challenging by adding zero-mean Gaussian noise to the ground-truth 2D detection on the Human3.6M dataset \cite{Human3.6M} (Fig.~\ref{fig:noise}). To ensure a fair comparison, we keep the input sequence length the same for all methods (in this case, 27 frames). For our method, $f$ = $n$ = 3. The experimental evidence reveals that PoseFormerV2 suffers from less performance drop as the standard deviation of Gaussian noise ($sigma$) increases while being more efficient. We observe that the performance of PoseFormerV1 drops drastically as $sigma$ increases from 8 to 10. In contrast, the proposed method presents a more stable trend. Moreover, our method even outperforms MHFormer \cite{Li_2022_CVPR} that incorporates the uncertainty of 2D detectors into the model design. Intriguingly, we find that minor noise may improve the accuracy of 3D pose estimation ($sigma$ = 3).
 
 \textbf{MPI-INF-3DHP.} We also compare our method with others on MPI-INF-3DHP \cite{MPIINF} (Table \ref{tab:mpi}). We use 9 central frames and the first 9 DCT coefficients from the input 81-frame sequence. The proposed method outperforms other approaches including P-STMO \cite{shan2022p} with masked joint pre-training. This result verifies the effectiveness of our method. Our implementation follows \cite{shan2022p}.

\begin{figure*}[t]
  \centering
    \includegraphics[width=1.0\linewidth]{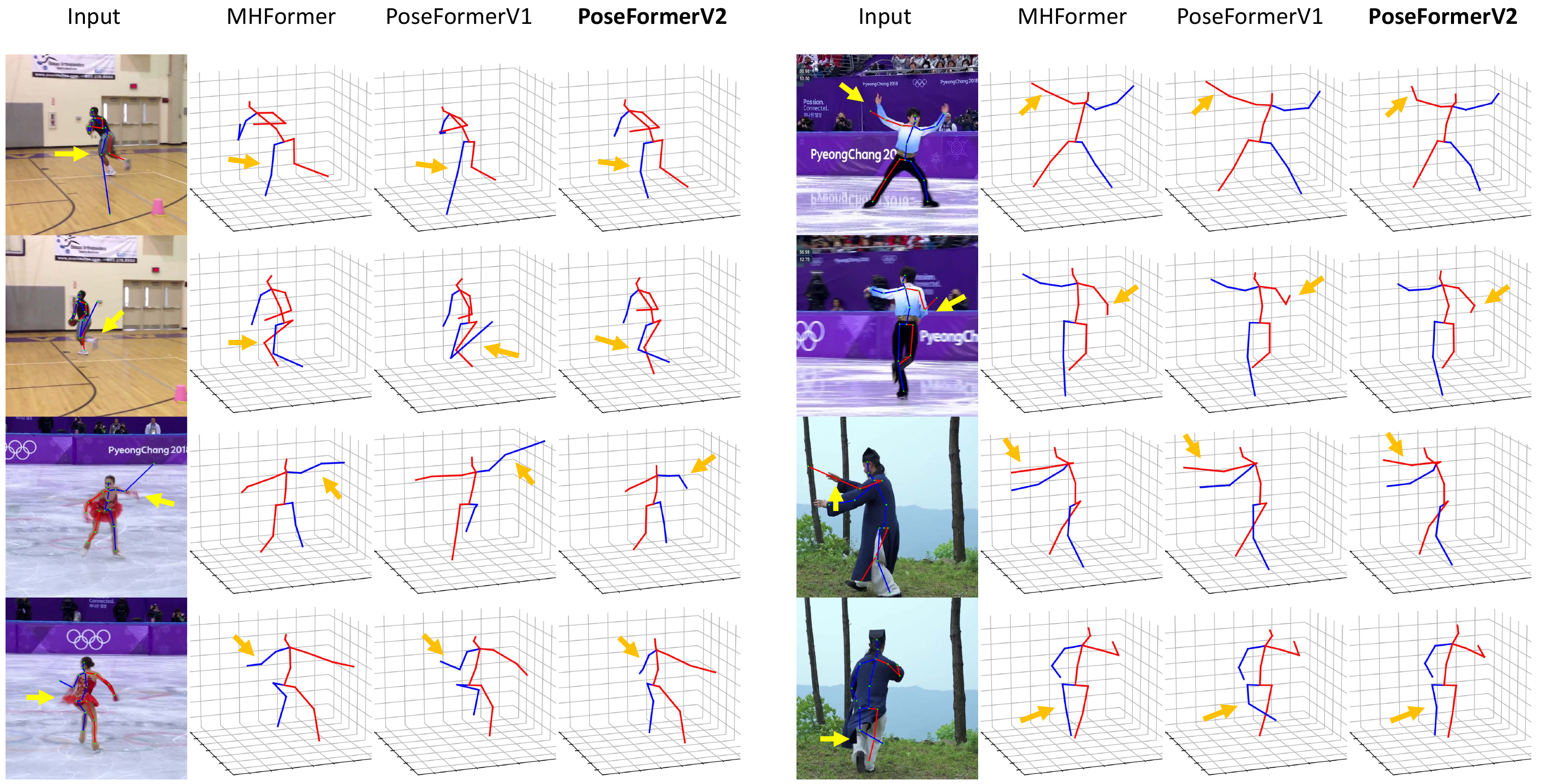}
    \caption{Qualitative comparisons of PoseFormerV2 with MHFormer \cite{Li_2022_CVPR} and PoseFormerV1 \cite{Zheng_2021_ICCV}. We randomly add Gaussian noise to the 2D detection of a specific joint. We highlight the deviated 2D detection with light-yellow arrows and corresponding 3D pose estimations with orange arrows. PoseFormerV2 shows better robustness to highly noisy input than existing methods.}
   \label{fig: qualitative comparisons}
\vspace{-10pt}
\end{figure*}

\textbf{Qualitative comparisons.} We provide qualitative comparisons of our method with competitive MHFormer \cite{Li_2022_CVPR} and PoseFormerV1 \cite{Zheng_2021_ICCV} in Fig.~\ref{fig: qualitative comparisons}. All methods use 81-frame 2D joint sequences as input. To further illustrate the robustness of our approach, we make the pose estimation task more difficult by adding Gaussian noise to the sequential 2D detection of a randomly selected joint (\eg, ``left wrist'', ``right foot''). The proposed method obtains reliable 3D human pose even under highly-deviated 2D detection (indicated by the light-yellow arrows). Note that our model is $\sim$9$\times$ more efficient than MHFormer (3.12 GFLOPs \vs 0.35 GFLOPs) and $\sim$4$\times$ more efficient compared to PoseFormerV1 (1.36 GFLOPs \vs 0.35 GFLOPs).
 
 \begin{table}[]
\scriptsize
\centering
\caption{Ablation study on several modifications to PoseFormerV1. We show how a 9-frame PoseFormerV1 is converted to PoseFormerV2 (with 9 DCT coefficients from an 81-frame sequence) step by step. The evaluation is performed on Human3.6M (Protocol 1, in mm). RF indicates Receptive Field.}
{\begin{tabular}{c|c|c|c} 
\hline
\begin{tabular}[c]{@{}c@{}} Step \end{tabular} &
\begin{tabular}[c]{@{}c@{}} Description \end{tabular} &  
\begin{tabular}[c]{@{}c@{}} RF \end{tabular} & 
\begin{tabular}[c]{@{}c@{}} MPJPE $\downarrow$ \end{tabular}  \\ 
\hline
(0) & Original 9-frame PoseFormerV1. & 9 & 49.9 \\
(1) & Frames are sampled from a longer sequence. & 9 & 49.9   \\
(2) & Append the embedding of DCT coefficients. & 81 & 47.1 (2.8$\downarrow$) \\
(3) & Replace the vanilla MLP with FreqMLP. & 81 & 46.0 (3.9$\downarrow$) \\
\hline
\end{tabular}}
\label{tab:ab1}
\vspace{-5pt}
\end{table}

 \begin{table}[]
\scriptsize
\centering
\caption{Ablation study on the number of frames and the number of DCT coefficients that are used as input to PoseFormerV2. The evaluation is performed on Human3.6M (Protocol 1, in mm).}
  {\begin{tabular}{c|c|c|c|c} 
\hline
\begin{tabular}[c]{@{}c@{}} Frame \\Number ($f$) \end{tabular} &
\begin{tabular}[c]{@{}c@{}} Coefficient \\Number ($n$) \end{tabular} & \begin{tabular}[c]{@{}c@{}} Full \\Length \end{tabular} &
\begin{tabular}[c]{@{}c@{}} MFLOPs \end{tabular} & MPJPE  \\ 
\hline
1 & 1 & 27 & 39.2 & 51.1   \\
1 & 3 & 27 & 77.2 & 48.7 (2.4$\downarrow$)    \\
3 & 1 & 27 & 79.4 & 50.1 (1.0$\downarrow$)    \\
3 & 3 & 27 & 117.3 & 47.9 (3.2$\downarrow$)    \\
9 & 9 & 27 & 351.7 & 47.6 (3.5$\downarrow$)    \\
\hline
\end{tabular}}
\label{tab:ab2}
\vspace{-10pt}
\end{table}

 \subsection{Ablation Study}
 \label{sec:ablation}
 

In this section, we show how a few modifications to PoseFormerV1 bring significant improvements in a step-by-step way. Moreover, to investigate more insights into the frequency-domain representation of input sequences, we reveal the impact of the number of input frames and that of kept DCT coefficients on our method.
 
 \textbf{Convert PoseFormerV1 into PoseFormerV2.} We inherit the overall spatial-temporal architecture from PoseFormerV1 and introduce restrained modifications to its temporal transformer for better multi-domain feature fusion. To exemplify, we illustrate how a 9-frame PoseFormerV1 is converted to PoseFormerV2 step by step: \textbf{(1)} the input (\ie, 9 frames) is sampled from a longer sequence (\eg, 81 frames) at the sequence center. This step brings \emph{no} performance improvement or increase in the receptive field since the input to the model is in fact unchanged. \textbf{(2)} The output of the spatial encoder of PoseFormerV1, $\mathbf{z}^{Time}$, is appended to the embedding of the first $n$ DCT coefficients (denoted by $\mathbf{z}^{Freq}$) of the complete sequence (81 frames in this case) as input into the temporal encoder. For convenience, we set $n$ to 9. \textbf{(3)} We replace the vanilla MLP for $\mathbf{z}^{Time}$ ($\mathbf{z}^{Time}$ and $\mathbf{z}^{Freq}$ already use separate vanilla MLPs before replacement) in the temporal encoder with FreqMLP (details in Sec. \ref{sec:PoseFormerV2}). PoseFormerV1 is converted to PoseFormerV2 after these steps, with an enlarged receptive field (from 9 to 81). We present the improvement brought by each step in Table \ref{tab:ab1}. It is worth noting that by introducing 9 DCT coefficients from a longer sequence (\ie, 81 frames), the MPJPE of 9-frame PoseFormerV1 is reduced by \textbf{7.8\%} (49.9mm \vs 46.0mm), which verifies the effectiveness of the proposed DCT representation of input joint sequences.
 
 \textbf{Number of input frames and DCT coefficients.} 
 In Table \ref{tab:ab2}, we investigate the impact of the number of frames ($f$) as input to the spatial encoder and the number of retained DCT coefficients ($n$). Here we keep the length of the entire joint sequence fixed, \ie, 27. The baseline model uses only one central frame and one DCT coefficient ($f$ = $n$ = 1). Increasing both parameters brings consistent improvements, and the increase in $n$ translates to more error reduction (2.4$\downarrow$ for $n$ = 3 \vs 1.0$\downarrow$ for $f$ = 3) since only a few DCT coefficients help capture the global characteristics of the entire sequence. We empirically find that the matched $f$ and $n$ with an expanding ratio of 9 (\ie, $f$ = $n$ = 3) achieve a satisfactory speed-accuracy trade-off.

 \subsection{Generalization Ability}
The proposed frequency-domain approach can generalize to other methods, \eg, MixSTE \cite{zhang2022mixste} and MHFormer \cite{Li_2022_CVPR}, as they also use transformers for temporal modeling. We improve both methods by incorporating low-frequency DCT coefficients. Details are in \textcolor{blue}{supplementary}. 
 
\section{Conclusion}
\label{sec:conclusion}
We present a solution to reconcile two seemingly unrelated or even contracted issues in lifting-based 3D human pose estimation -- the efficiency of processing long-sequence input and the robustness to noisy joint detection -- simultaneously from a barely explored frequency-domain perspective. The proposed method, PoseFormerV2, exploits a compact frequency representation of long 2D joint sequences to efficiently enlarge the receptive field of the model while improving its robustness. Experimental results show that our method outperforms previous transformer-based methods on Human3.6M and MPI-INF-3DHP. 
 
\appendix
\section*{Supplementary Material}

\section{Overview}
The supplementary material includes sections as follows:
\begin{itemize}
\item Section \ref{sec: DCT}: A formal introduction to Discrete Cosine Transform.

\item Section \ref{sec: datasets}: Datasets and evaluation metrics.

\item Section \ref{sec: details}: More implementation details.

\item Section \ref{sec: baseline}: Comparisons of PoseFormerV2 and a simple baseline model purely in the frequency domain.

\item Section \ref{sec: generalization}: Generalization of our approach to more models.

\item Section \ref{sec: visualization}: Visualizations and analysis.

\item Section \ref{sec: impact}: Broader impacts and limitations.
\end{itemize}

\section{Discrete Cosine Transform}
\label{sec: DCT}
We now give a formal introduction to DCT. Given a 2D joint sequence denoted by $\mathbf{x} \in \mathbb{R} ^{F \times J \times 2}$, where ${F}$ is the sequence length and ${J}$ is the joint number in each frame, the trajectory of the x (or y) coordinate of the j-{th} joint denoted as $\mathbf{x}_{j,0} \in \mathbb{R} ^{F}$ (or $\mathbf{x}_{j,1} \in \mathbb{R} ^{F}$, both denoted by $\hat{\mathbf{x}}_j$ for convenience) is a 1D time series and we apply DCT to each trajectory ($J*2$ trajectories in total) individually.

For trajectory $\hat{\mathbf{x}}_j$, the $i$-{th} DCT coefficient is calculated as
\vspace{-0.2cm}
\begin{equation}
    \resizebox{0.9\linewidth}{!}{$C_{j,i} = \sqrt{\frac{2}{F}}\sum_{f=1}^{F}x_{j,f}\frac{1}{\sqrt{1+\delta_{i1}}}\cos\left(\frac{\pi}{2F}(2f-1)(i-1)\right)\;,$}
    \label{eq:dct}
    \vspace{-0.2cm}
\end{equation}
where $\delta_{i1} = 1\ when\ i=1,\ otherwise\ \delta_{i1} = 0$. Each time step in trajectory yields one DCT coefficient, \ie, $i\in\{1,2,\cdots,F\}$. DCT coefficients encode multiple levels of temporal information in the input time series. Specifically, low-frequency coefficients (\ie, when $i$ is small) encode the rough contour of the input sequence while high-frequency coefficients (\ie, for the large $i$) encode details, \eg, jitters or sharp changes in the input sequence. The original input sequence in the time domain can be restored using Inverse Discrete Cosine Transform (IDCT), which is given by
\vspace{-0.2cm}
\begin{equation}
    \resizebox{0.9\linewidth}{!}{$x_{j,f} =\sqrt{\frac{2}{F}}\sum_{i=1}^{F}C_{j,i}\frac{1}{\sqrt{1+\delta_{i1}}}\cos\left(\frac{\pi}{2F}(2f-1)(i-1)\right)\;,$}
    \vspace{-0.2cm}
\end{equation}
and $f\in\{1,2,\cdots,F\}$. DCT is lossless if we keep all its coefficients intact. In practice, we can slightly lossily recover the input sequence using only a few low-frequency coefficients and set other coefficients to zero. It is worth noting that the recovered curve would be smoother compared to the original one since we discard some of the high-frequency coefficients.
This property of DCT is desirable -- only a small proportion of DCT coefficients are enough to represent the whole input sequence, even in a cleaner manner. This motivates us to use such representation to efficiently operate long sequences while improving the robustness of the model to low-quality 2D detection where high-frequency noise often occurs.

\section{Datasets and Evaluation Metrics}
\label{sec: datasets}
\textbf{Human3.6M} is the most widely used benchmark for 3D human pose estimation. Over 3.6 million video frames are captured indoors from 4 cameras at different places. This dataset contains 11 subjects performing 15 different actions, \eg, “Walking” and “Phoning”. We train our model on 5 subjects (S1, S5, S6, S7, S8) and use other 2 subjects (S9, S11) for testing, following \cite{pavllo2019,Liu_2020_CVPR,chen2020anatomy, Zheng_2021_ICCV}.

\textbf{MPI-INF-3DHP} is collected in both controlled indoor environments and challenging outdoor environments. It also provides different subjects and actions from multiple camera views similar to Human3.6M.

\textbf{Evaluation Metrics}. We report two common metrics, MPJPE and P-MPJPE \cite{zheng2020deep} on Human3.6M. MPJPE (Mean Per Joint Position Error, referred to as Protocol 1) measures the mean Euclidean distance between the estimated 3D pose and the ground truth 3D pose. P-MPJPE (Protocol 2) applies a rigid transformation to the estimated 3D pose and the distance is computed between the aligned estimated 3D pose and the ground truth 3D pose.

For the MPI-INF-3DHP dataset, we report MPJPE, Percentage of Correct Keypoint (PCK) within the 150mm range, and Area Under Curve (AUC) as in \cite{lin2019trajectory,chen2020anatomy,wang2020motion}.

\section{More Implementation Details}
\label{sec: details}
 Our method is built upon PoseFormerV1 \cite{Zheng_2021_ICCV}. Aiming at better demonstrating the effectiveness of our DCT coefficient representation of input sequences and providing fair comparisons to PoseFormerV1, we \textbf{directly} adopt optimal hyper-parameters for model architecture from PoseFormerV1, although further investigation may bring additional improvements. 
 
\textbf{Model architecture hyper-parameters}. The embedded feature dimension $c$ in the spatial transformer is 32 and the layer number of the spatial transformer and feature-fusion transformer is 4, following \cite{Zheng_2021_ICCV}. Plus, the design of Spatial-Temporal Positional Embedding is also adopted from \cite{Zheng_2021_ICCV}.
 
 \textbf{Experimental settings}. Our experiments are conducted with Pytorch \cite{PyTorch} on a single NVIDIA RTX 3090. For both training and testing, we apply horizontal flipping augmentation following \cite{pavllo2019, Liu_2020_CVPR, chen2020anatomy, Zheng_2021_ICCV}. We train our model using the AdamW \cite{loshchilov2017decoupled} optimizer for 80 epochs with a weight decay of 0.1. The initial learning rate is set to 8e-4 with an exponential learning rate decay schedule and the decay factor is 0.99. We adopt the CPN \cite{chen2018cpn} 2D pose detection on Human3.6M, following \cite{pavllo2019,Liu_2020_CVPR,chen2020anatomy}. As for the MPI-INF-3DHP dataset, we use ground truth 2D detection, following \cite{mehta2017vnect,lin2019trajectory}. 
 
\section{Simple Baseline}
\label{sec: baseline}
 In our approach, the temporal encoder of PoseFormerV1 \cite{Zheng_2021_ICCV} is reformulated as a Time-Frequency Feature Fusion module and we show that the low-frequency coefficients of the input sequence help improve the efficiency of the model to process long sequences and its robustness against noisy joint detection. Given the effectiveness of this representation, readers may raise a question: Why not entirely extract features from DCT coefficients of the input sequence but additionally combine them with features in the time domain? Here we design a baseline model where we simply replace the input to PoseFormerV1 \cite{Zheng_2021_ICCV} (joint coordinates in the time domain) with low-frequency DCT coefficients of the input sequence. The full sequence length and the number of the retained DCT coefficients (denoted as $n$) are kept the same for the baseline model and our approach. For convenience, the number of frames ($f$) as input into the spatial encoder of PoseFormerV2 is set to $n$. We provide quantitative results to demonstrate that this straightforward approach does not work well, especially when the ratio between the full sequence length and $n$ is increased (see Table \ref{tab:baseline}). The features of only a few central frames in the sequence significantly boost accuracy, \eg, with 3 central frames of the full input sequence of length 81, the MPJPE is reduced from 49.7mm to 47.1mm (5.2\%$\downarrow$, the 3rd row in Table \ref{tab:baseline}). 
 
 Intuitively, the spatial encoder of PoseFormerV2 that encodes joint coordinates of a few central frames in the time domain helps capture the fine-grained human motions, benefiting 3D pose estimation for the frame at the sequence center. In contrast, low-frequency coefficients of the input sequence filter out high-frequency noise and human motion details (\eg, fast motions) that may be informative to human pose estimation (\ie, the over-smoothing problem). Therefore, features from the time domain and frequency domain, \ie, the joint coordinate of central frames and low-frequency coefficients of the sequence, carry complementary semantics. These considerations necessitate our proposed Time-Frequency Feature Fusion design.

 \begin{table}[]
\scriptsize
\centering
\caption{Comparisons of PoseFormerV2 and a simple baseline. The evaluation is performed on Human3.6M (Protocol 1, MPJPE) \cite{Human3.6M} and the Frame Number ($f$) is only applicable to PoseFormerV2.}
  {\begin{tabular}{c|c|c|c|c} 
\hline
\begin{tabular}[c]{@{}c@{}} Frame \\Number ($f$) \end{tabular} &
\begin{tabular}[c]{@{}c@{}} Coefficient \\Number ($n$) \end{tabular} & \begin{tabular}[c]{@{}c@{}} Full \\Length \end{tabular} &
\begin{tabular}[c]{@{}c@{}} Baseline \end{tabular} & PoseFormerV2  \\ 
\hline
3 & 3 & 9 & 50.2 & 49.5 (0.7$\downarrow$)   \\
3 & 3 & 27 & 48.7 & 47.9 (0.8$\downarrow$)    \\
3 & 3 & 81 & 49.7 & 47.1 (2.6$\downarrow$)    \\
9 & 9 & 27 & 48.8 & 47.6 (1.2$\downarrow$)    \\
9 & 9 & 81 & 47.8 & 46.0 (1.8$\downarrow$)    \\
\hline
\end{tabular}}
\label{tab:baseline}
\vspace{-5pt}
\end{table}

\section{Generalization to More Models}
\label{sec: generalization}
In the main text, we focus on improving PoseFormerV1 \cite{Zheng_2021_ICCV} from a barely explored frequency-domain perspective. In this part, we show that the proposed frequency-domain approach also generalizes well to other existing state-of-the-art methods, \eg, MixSTE \cite{zhang2022mixste} and MHFormer \cite{Li_2022_CVPR}. Since these approaches \cite{Li_2022_CVPR, zhang2022mixste} also apply self-attention along the time dimension to all frames as PoseFormerV1 \cite{Zheng_2021_ICCV}, the proposed method can be easily incorporated into their model without complex redesigns for model architecture. For fair comparisons, we \textbf{directly} adopt optimal hyper-parameters (\eg, the layer number, channel dimension) for these original methods. Further tuning of hyper-parameters may bring additional improvements.

\begin{figure}[t]
  \centering
   \includegraphics[width=0.96\linewidth]{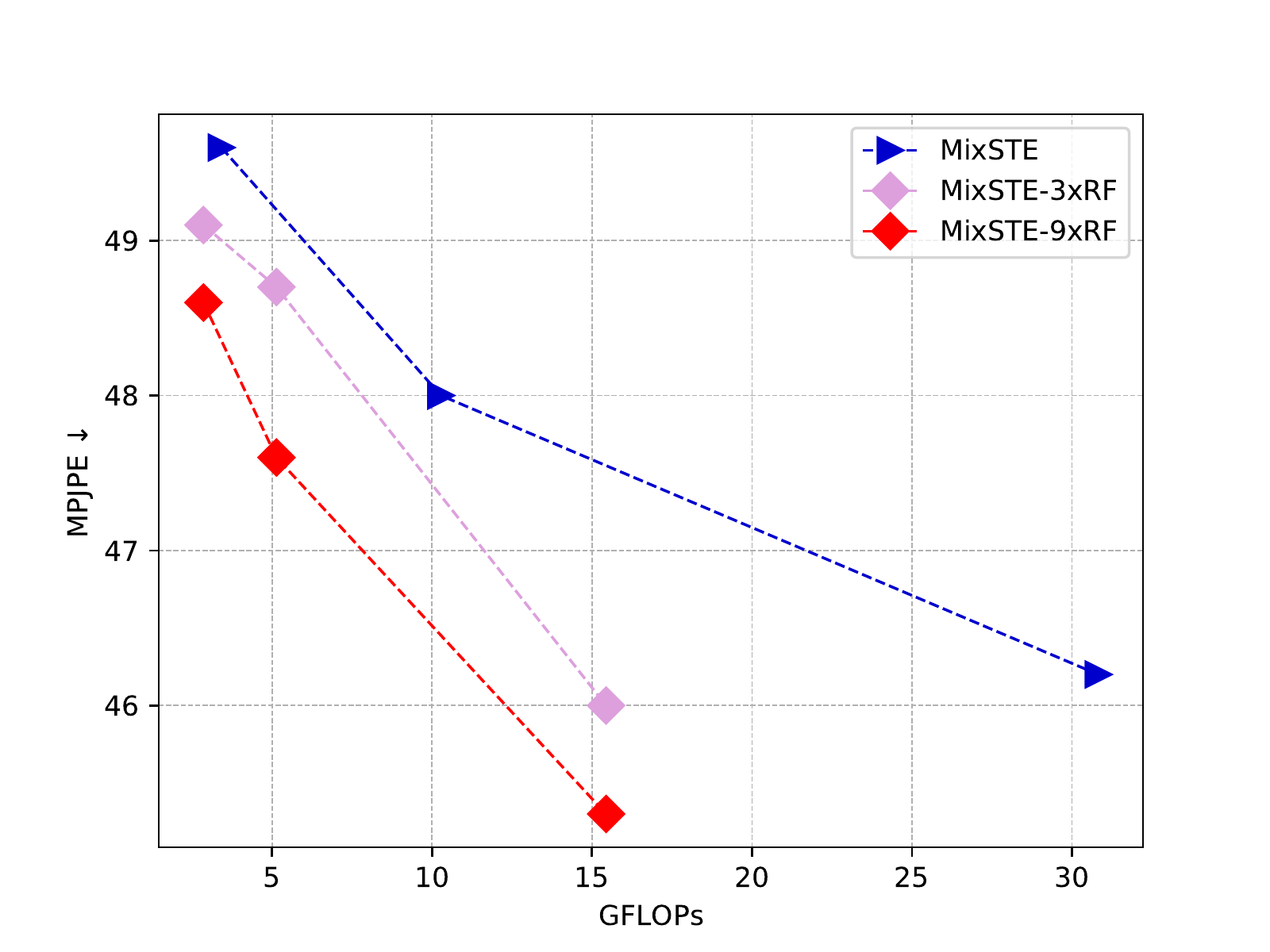}
   \vspace{-3pt}
   \caption{Comparisons of MixSTE \cite{zhang2022mixste} and its improved version with frequency representations of the sequence on Human3.6M \cite{Human3.6M}. RF: Receptive Field and $k\times$RF indicate that the RF of MixSTE is expanded by $k\times$ with a few low-frequency DCT coefficients of the full sequence. The proposed approach helps MixSTE gain a better speed-accuracy trade-off. (Best viewed in color)}
   \label{fig: Human3.6M_Mix}
   \vspace{-5pt}
\end{figure}

\begin{figure}[t]
\vspace{0pt}
  \centering
   \includegraphics[width=0.96\linewidth]{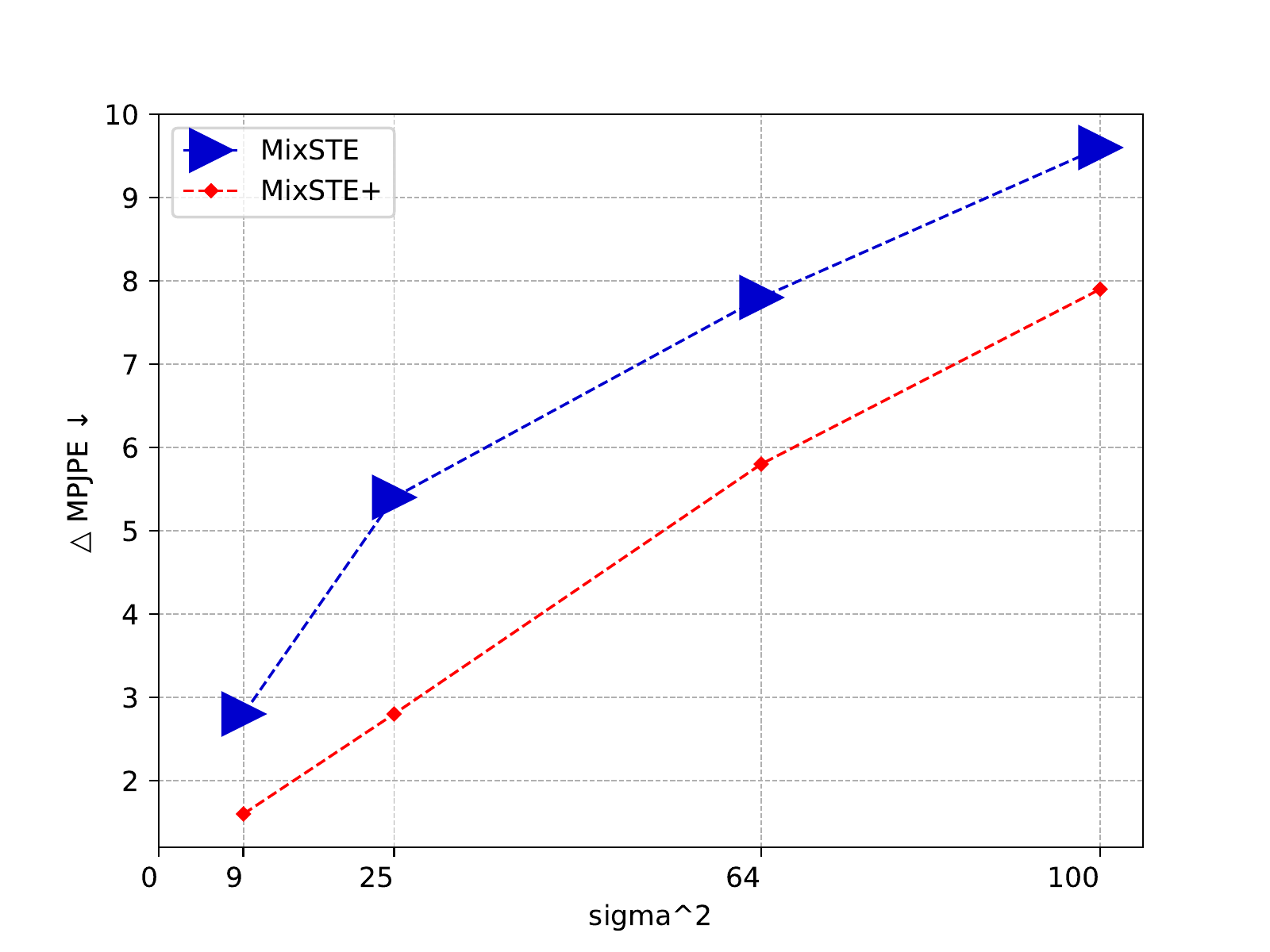}
   \caption{Comparisons of MixSTE \cite{zhang2022mixste} and its improved version using low-frequency DCT coefficients of the sequence in terms of robustness to noise on Human3.6M \cite{Human3.6M}. Zero-mean Gaussian noise of standard deviation $sigma$ is added to ground truth 2D detection, and we show their performance drop ($\triangle $MPJPE) as $sigma$ increases. \textcolor{blue}{The size of markers indicates the computational cost of models.}}
   \vspace{-5pt}
   \label{fig: noise_Mix}
\end{figure}

\begin{figure}[t]
\vspace{-5pt}
  \centering
   \includegraphics[width=0.96\linewidth]{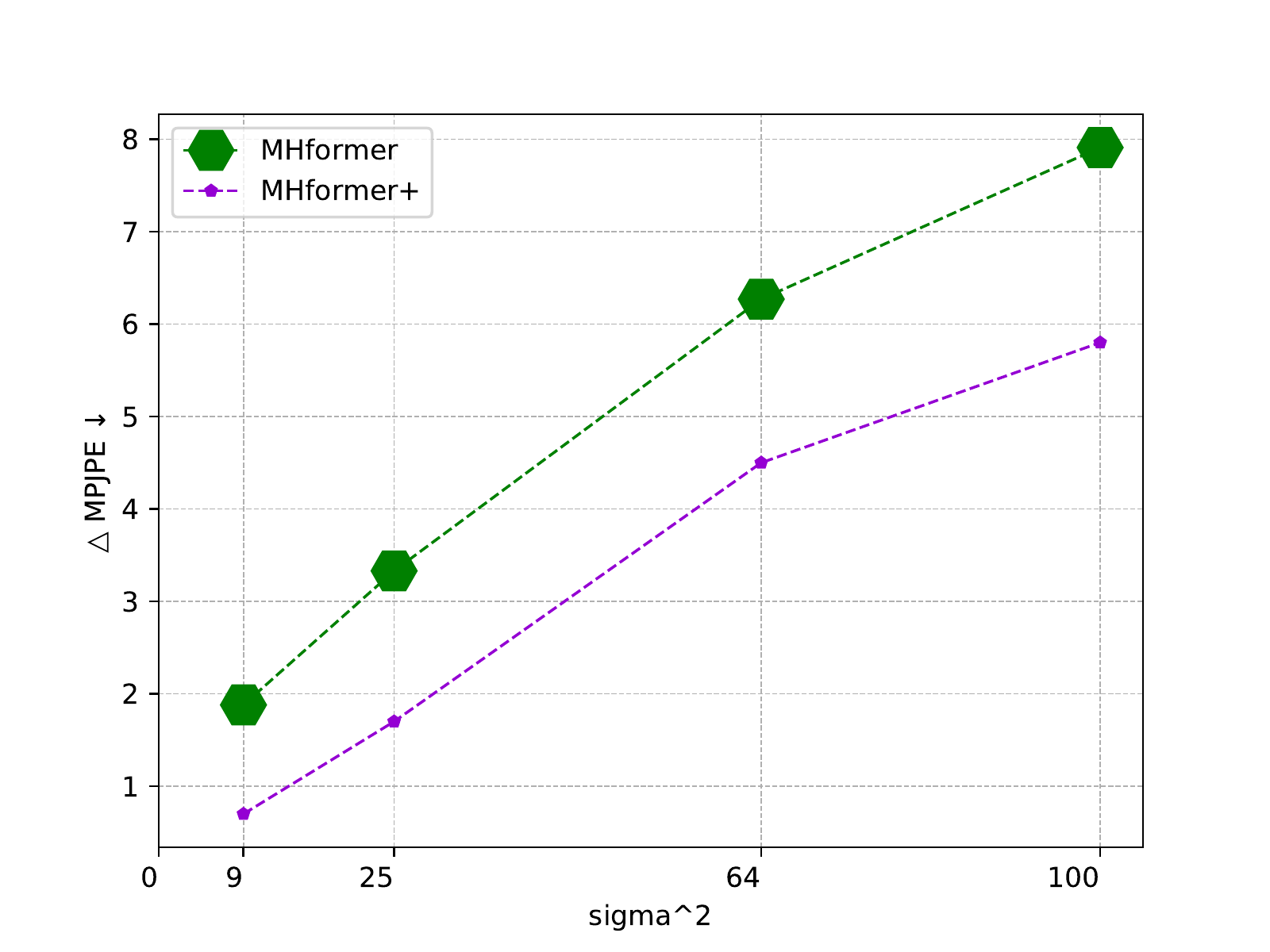}
   \caption{Comparisons of MHFormer \cite{Li_2022_CVPR} and its improved version using low-frequency DCT coefficients of the sequence in terms of robustness to noise on Human3.6M \cite{Human3.6M}. Experimental settings follow Fig. \ref{fig: noise_Mix}. \textcolor{blue}{The size of markers indicates the computational cost of models.}}
   \vspace{-5pt}
   \label{fig: noise_MH}
\end{figure}

\textbf{MixSTE \cite{zhang2022mixste}} adopts the spatial-temporal architecture as PoseFormerV1 \cite{Zheng_2021_ICCV}. Compared to the spatial-then-temporal paradigm of PoseFormerV1, MixSTE alternately uses spatial and temporal transformer encoders. Similarly, we centrally sample a few video frames from a longer sequence as input into the spatial encoders of MixSTE. For temporal encoders, we append the time-domain features (the output of the spatial encoders) with the embedding of low-frequency coefficients of the complete input sequence. The comparisons of MixSTE and its improved version are presented in Fig.~\ref{fig: Human3.6M_Mix}, \ref{fig: noise_Mix}. Original MixSTE is highly computationally expensive and our approach improves its efficiency and accuracy simultaneously, \eg, MixSTE achieves 46.2mm MPJPE taking 30.8 GFLOPs, while its improved version achieves 45.3mm MPJPE with 15.4 GFLOPs (2$\times$ faster and 1.9\%$\uparrow$ error reduction, see the bright red curve in Fig.~\ref{fig: Human3.6M_Mix}). We also show that our method improves the robustness of MixSTE against noisy 2D joint detection (Fig.~\ref{fig: noise_Mix}). Specifically, we add zero-mean Gaussian noise to the ground-truth 2D joint sequence of 27 frames on Human3.6M \cite{Human3.6M}. The improved MixSTE (denoted as MixSTE+) uses 3 central frames as input to its spatial encoders and the first 3 DCT coefficients as a cleaner global representation of the full sequence. MixSTE+ suffers from less performance drop while being 6$\times$ more efficient (30.8 GFLOPs \vs 5.1 GFLOPs, indicated by the marker size).

\textbf{MHFormer \cite{Li_2022_CVPR}} introduces multiple hypotheses into its architecture to model depth ambiguity of body parts and uncertainty of joint detectors and is thus relatively robust (experimental results are available in the main paper). Besides, MHFormer also includes spatial-temporal transformer modules as in PoseFormerV1 \cite{Zheng_2021_ICCV}. To further verify the universality of our approach, we similarly improve MHFormer following MixSTE+. Experimental evidence shows that the proposed method promotes the robustness of MHFormer while reducing its computational cost (see Fig.~\ref{fig: noise_MH}), even though it already equips itself with prior knowledge of noisy joint detection. Therefore, this result demonstrates that, in terms of improvements in the robustness of models, our method is compatible with other approaches.

We have so far generalized our approach to other two transformer-based methods, \ie, MixSTE \cite{zhang2022mixste} and MHFormer \cite{Li_2022_CVPR}. We may explore the generalization of the proposed method to a wider range of model architectures in the future, such as CNN-based and GNN-based methods in 3D human pose estimation. Moreover, we believe our method can also be utilized in other tasks, especially skeleton-based ones where the computational cost of long-sequence processing and the quality of human skeleton representations can become problems.

\begin{figure*}[t]
  \centering
   \includegraphics[width=1.0\linewidth]{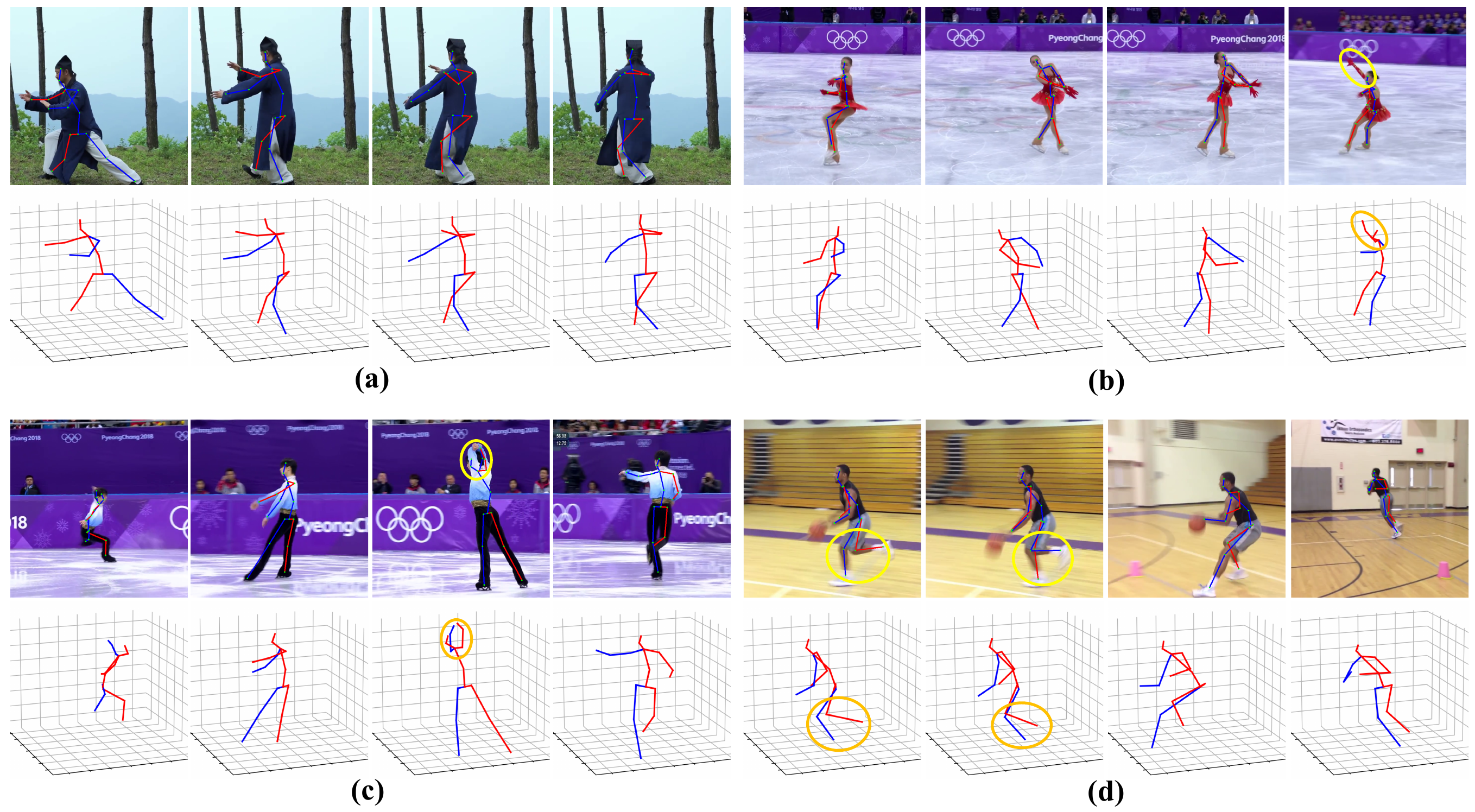}
    \caption{Qualitative results of PoseFormerV2 under challenging in-the-wild images: (a) Occlusions; (b)(c) Missed 2D joint detection; (d) Switched 2D joints. We highlight the unreliable 2D detection with light-yellow circles and corresponding 3D pose estimations with orange circles. PoseFormerV2 shows great robustness to imperfect 2D joint detection.}
   \label{fig: qualitative results}
\end{figure*}


\section{Visualizations and Analysis}
\label{sec: visualization}
In this section, we provide a series of qualitative results on challenging in-the-wild images to showcase the robustness of PoseFormerV2 in real scenarios. 

Fig.~\ref{fig: qualitative results} presents several representative hard cases with HRNet \cite{sun2019deep} 2D joint detection: (a) Occlusions where joints overlap with each other; (b)(c) Missed joints; (d) Switched joints. Specifically, the right arm of the person in the 4th image of (b) and the left arm of the person in the 3rd image of (c) are missed. Moreover, in the 2nd image of (d), two legs of the person are switched (highlighted with light-yellow circles). Despite the imperfect 2D joint input, PoseFormerV2 still infers correct positions for these joints in 3D space (marked with orange circles).


\textbf{Analysis.} 
The robustness of PoseFormerV2 is attributed to the usage of an appropriate representation -- low-frequency DCT coefficients -- of the input joint sequence, instead of hand-crafted modules that may bring additional computational cost such as the multi-hypothesis generation module in \cite{Li_2022_CVPR}. Low-frequency DCT coefficients provide a global vision of the input sequence and therefore the noise contained in individual video frames is dwarfed. This utilization of DCT coefficients also brings an extra advantage to PoseFormerV2, the temporal consistency of the estimated 3D pose between adjacent frames. We provide a video demo to illustrate that the proposed method keeps an excellent consistency (\ie, temporal stability) under extremely corrupted 2D joint detection.

\section{Broader Impacts and Limitations}
\label{sec: impact}
\textbf{Broader impacts}. In this paper, we attempt to reconcile two critical issues in real-scenario applications of 3D HPE, \ie, the efficiency of models to process long sequences for improved precision and their robustness against noisy 2D detection as high-quality joint sequences are hard to obtain. To encourage more real-world applications, we may shift our research focus from marginal improvements on carefully controlled datasets to overcoming the drawbacks of existing approaches in practical use. We expect more research to follow this line.

On the other hand, this work is done based on a scarcely investigated frequency method, \ie, Discrete Cosine Transform (DCT) which plays an important role in conventional image compression algorithms. We hope this research will inspire more research to revisit traditional signal processing techniques as various data we treat in the deep learning era is actually signals of different forms. An appropriate combination of these choreographed techniques and recently developed deep learning approaches may bring surprising advantages.

\textbf{Limitations.} Our method includes two important hyper-parameters -- the number of sampled central frames and that of the kept DCT coefficients of the complete input sequence. Currently, they are chosen on the basis of experimental results or human experience for the trade-off between speed and accuracy. In the future, we may reshape them as learnable parameters that can be automatically learned from input data, or we may further theoretically formulate the optimal choices for them, thus removing the need for parameter-searching.

{\small
\bibliographystyle{ieee_fullname}
\bibliography{egbib}

\begin{thebibliography}{10}\itemsep=-1pt

\bibitem{Arnab_2021_ICCV}
Anurag Arnab, Mostafa Dehghani, Georg Heigold, Chen Sun, Mario Lu\v{c}i\'c, and
  Cordelia Schmid.
\newblock Vivit: A video vision transformer.
\newblock In {\em Proceedings of the IEEE/CVF International Conference on
  Computer Vision (ICCV)}, pages 6836--6846, October 2021.

\bibitem{gberta_2021_ICML}
Gedas Bertasius, Heng Wang, and Lorenzo Torresani.
\newblock Is space-time attention all you need for video understanding?
\newblock In {\em Proceedings of the International Conference on Machine
  Learning (ICML)}, July 2021.

\bibitem{brown2020language}
Tom Brown, Benjamin Mann, Nick Ryder, Melanie Subbiah, Jared~D Kaplan, Prafulla
  Dhariwal, Arvind Neelakantan, Pranav Shyam, Girish Sastry, Amanda Askell,
  et~al.
\newblock Language models are few-shot learners.
\newblock {\em Advances in neural information processing systems},
  33:1877--1901, 2020.

\bibitem{carion2020end}
Nicolas Carion, Francisco Massa, Gabriel Synnaeve, Nicolas Usunier, Alexander
  Kirillov, and Sergey Zagoruyko.
\newblock End-to-end object detection with transformers.
\newblock In {\em European conference on computer vision}, pages 213--229.
  Springer, 2020.

\bibitem{chen2020anatomy}
Tianlang Chen, Chen Fang, Xiaohui Shen, Yiheng Zhu, Zhili Chen, and Jiebo Luo.
\newblock Anatomy-aware 3d human pose estimation with bone-based pose
  decomposition.
\newblock {\em IEEE Transactions on Circuits and Systems for Video Technology},
  2021.

\bibitem{chen2018cascaded}
Yilun Chen, Zhicheng Wang, Yuxiang Peng, Zhiqiang Zhang, Gang Yu, and Jian Sun.
\newblock Cascaded pyramid network for multi-person pose estimation.
\newblock In {\em CVPR}, 2018.

\bibitem{chen2018cpn}
Yilun Chen, Zhicheng Wang, Yuxiang Peng, Zhiqiang Zhang, Gang Yu, and Jian Sun.
\newblock Cascaded pyramid network for multi-person pose estimation.
\newblock In {\em Proceedings of the IEEE conference on computer vision and
  pattern recognition}, pages 7103--7112, 2018.

\bibitem{devlin2018bert}
Jacob Devlin, Ming-Wei Chang, Kenton Lee, and Kristina Toutanova.
\newblock Bert: Pre-training of deep bidirectional transformers for language
  understanding.
\newblock {\em arXiv preprint arXiv:1810.04805}, 2018.

\bibitem{Dosovitskiy2020ViT}
Alexey Dosovitskiy, Lucas Beyer, Alexander Kolesnikov, Dirk Weissenborn,
  Xiaohua Zhai, Thomas Unterthiner, Mostafa Dehghani, Matthias Minderer, Georg
  Heigold, Sylvain Gelly, Jakob Uszkoreit, and Neil Houlsby.
\newblock An image is worth 16x16 words: Transformers for image recognition at
  scale.
\newblock {\em ICLR}, 2021.

\bibitem{einfalt_up3dhpe_WACV23}
Moritz Einfalt, Katja Ludwig, and Rainer Lienhart.
\newblock Uplift and upsample: Efficient 3d human pose estimation with
  uplifting transformers.
\newblock In {\em Proceedings of the IEEE/CVF Winter Conference on Applications
  of Computer Vision (WACV)}, January 2023.

\bibitem{guibas2021adaptive}
John Guibas, Morteza Mardani, Zongyi Li, Andrew Tao, Anima Anandkumar, and
  Bryan Catanzaro.
\newblock Adaptive fourier neural operators: Efficient token mixers for
  transformers.
\newblock {\em arXiv preprint arXiv:2111.13587}, 2021.

\bibitem{he2022masked}
Kaiming He, Xinlei Chen, Saining Xie, Yanghao Li, Piotr Doll{\'a}r, and Ross
  Girshick.
\newblock Masked autoencoders are scalable vision learners.
\newblock In {\em Proceedings of the IEEE/CVF Conference on Computer Vision and
  Pattern Recognition}, pages 16000--16009, 2022.

\bibitem{Human3.6M}
C. {Ionescu}, D. {Papava}, V. {Olaru}, and C. {Sminchisescu}.
\newblock Human3.6m: Large scale datasets and predictive methods for 3d human
  sensing in natural environments.
\newblock {\em IEEE TPAMI}, 2014.

\bibitem{Li_2020_CVPR}
Shichao Li, Lei Ke, Kevin Pratama, Yu-Wing Tai, Chi-Keung Tang, and Kwang-Ting
  Cheng.
\newblock Cascaded deep monocular 3d human pose estimation with evolutionary
  training data.
\newblock In {\em Proceedings of the IEEE/CVF Conference on Computer Vision and
  Pattern Recognition (CVPR)}, June 2020.

\bibitem{li2022exploiting}
Wenhao Li, Hong Liu, Runwei Ding, Mengyuan Liu, Pichao Wang, and Wenming Yang.
\newblock Exploiting temporal contexts with strided transformer for 3d human
  pose estimation.
\newblock {\em IEEE Transactions on Multimedia}, 2022.

\bibitem{Li_2022_CVPR}
Wenhao Li, Hong Liu, Hao Tang, Pichao Wang, and Luc Van~Gool.
\newblock Mhformer: Multi-hypothesis transformer for 3d human pose estimation.
\newblock In {\em Proceedings of the IEEE/CVF Conference on Computer Vision and
  Pattern Recognition (CVPR)}, pages 13147--13156, June 2022.

\bibitem{lin2019trajectory}
Jiahao Lin and Gim~Hee Lee.
\newblock Trajectory space factorization for deep video-based 3d human pose
  estimation.
\newblock In {\em BMVC}, 2019.

\bibitem{Liu_2020_CVPR}
Ruixu Liu, Ju Shen, He Wang, Chen Chen, Sen-ching Cheung, and Vijayan Asari.
\newblock Attention mechanism exploits temporal contexts: Real-time 3d human
  pose reconstruction.
\newblock In {\em CVPR}, 2020.

\bibitem{liu2021swin}
Ze Liu, Yutong Lin, Yue Cao, Han Hu, Yixuan Wei, Zheng Zhang, Stephen Lin, and
  Baining Guo.
\newblock Swin transformer: Hierarchical vision transformer using shifted
  windows.
\newblock In {\em Proceedings of the IEEE/CVF International Conference on
  Computer Vision}, pages 10012--10022, 2021.

\bibitem{loshchilov2017decoupled}
Ilya Loshchilov and Frank Hutter.
\newblock Decoupled weight decay regularization.
\newblock {\em arXiv preprint arXiv:1711.05101}, 2017.

\bibitem{mao2020history}
Wei Mao, Miaomiao Liu, and Mathieu Salzmann.
\newblock History repeats itself: Human motion prediction via motion attention.
\newblock In {\em European Conference on Computer Vision}, pages 474--489.
  Springer, 2020.

\bibitem{mao2019learning}
Wei Mao, Miaomiao Liu, Mathieu Salzmann, and Hongdong Li.
\newblock Learning trajectory dependencies for human motion prediction.
\newblock In {\em Proceedings of the IEEE/CVF International Conference on
  Computer Vision}, pages 9489--9497, 2019.

\bibitem{MPIINF}
Dushyant Mehta, Helge Rhodin, Dan Casas, Pascal Fua, Oleksandr Sotnychenko,
  Weipeng Xu, and Christian Theobalt.
\newblock Monocular 3d human pose estimation in the wild using improved cnn
  supervision.
\newblock In {\em 2017 international conference on 3D vision (3DV)}, pages
  506--516. IEEE, 2017.

\bibitem{mehta2017vnect}
Dushyant Mehta, Srinath Sridhar, Oleksandr Sotnychenko, Helge Rhodin, Mohammad
  Shafiei, Hans-Peter Seidel, Weipeng Xu, Dan Casas, and Christian Theobalt.
\newblock Vnect: Real-time 3d human pose estimation with a single rgb camera.
\newblock {\em ACM Transactions on Graphics (TOG)}, 36(4):1--14, 2017.

\bibitem{Moon_I2L_MeshNet}
Gyeongsik Moon and Kyoung~Mu Lee.
\newblock I2l-meshnet: Image-to-lixel prediction network for accurate 3d human
  pose and mesh estimation from a single rgb image.
\newblock In {\em ECCV}, 2020.

\bibitem{newell2016stacked}
Alejandro Newell, Kaiyu Yang, and Jia Deng.
\newblock Stacked hourglass networks for human pose estimation.
\newblock In {\em European conference on computer vision}, pages 483--499.
  Springer, 2016.

\bibitem{PyTorch}
Adam Paszke, Sam Gross, Soumith Chintala, Gregory Chanan, Edward Yang, Zachary
  DeVito, Zeming Lin, Alban Desmaison, Luca Antiga, and Adam Lerer.
\newblock Automatic differentiation in pytorch.
\newblock 2017.

\bibitem{pavlakos2018ordinal}
Georgios Pavlakos, Xiaowei Zhou, and Kostas Daniilidis.
\newblock Ordinal depth supervision for 3{D} human pose estimation.
\newblock In {\em CVPR}, 2018.

\bibitem{pavllo2019}
Dario Pavllo, Christoph Feichtenhofer, David Grangier, and Michael Auli.
\newblock 3d human pose estimation in video with temporal convolutions and
  semi-supervised training.
\newblock In {\em CVPR}, 2019.

\bibitem{pennebaker1992jpeg}
William~B Pennebaker and Joan~L Mitchell.
\newblock {\em JPEG: Still image data compression standard}.
\newblock Springer Science \& Business Media, 1992.

\bibitem{rao2021global}
Yongming Rao, Wenliang Zhao, Zheng Zhu, Jiwen Lu, and Jie Zhou.
\newblock Global filter networks for image classification.
\newblock {\em Advances in Neural Information Processing Systems}, 34:980--993,
  2021.

\bibitem{shan2022p}
Wenkang Shan, Zhenhua Liu, Xinfeng Zhang, Shanshe Wang, Siwei Ma, and Wen Gao.
\newblock P-stmo: Pre-trained spatial temporal many-to-one model for 3d human
  pose estimation.
\newblock {\em arXiv preprint arXiv:2203.07628}, 2022.

\bibitem{skodras2001jpeg}
Athanassios Skodras, Charilaos Christopoulos, and Touradj Ebrahimi.
\newblock The jpeg 2000 still image compression standard.
\newblock {\em IEEE Signal processing magazine}, 18(5):36--58, 2001.

\bibitem{sun2019deep}
Ke Sun, Bin Xiao, Dong Liu, and Jingdong Wang.
\newblock Deep high-resolution representation learning for human pose
  estimation.
\newblock In {\em CVPR}, 2019.

\bibitem{touvron2020deit}
Hugo Touvron, Matthieu Cord, Matthijs Douze, Francisco Massa, Alexandre
  Sablayrolles, and Herv\'e J\'egou.
\newblock Training data-efficient image transformers \& distillation through
  attention.
\newblock {\em arXiv preprint arXiv:2012.12877}, 2020.

\bibitem{Attention_is_All_You_Need}
Ashish Vaswani, Noam Shazeer, Niki Parmar, Jakob Uszkoreit, Llion Jones,
  Aidan~N. Gomez, Lukasz Kaiser, and Illia Polosukhin.
\newblock Attention is all you need.
\newblock 2017.

\bibitem{wang2020motion}
Jingbo Wang, Sijie Yan, Yuanjun Xiong, and Dahua Lin.
\newblock Motion guided 3d pose estimation from videos.
\newblock In {\em European Conference on Computer Vision}, pages 764--780.
  Springer, 2020.

\bibitem{wang2022vtclfc}
Zhenyu Wang, Hao Luo, Pichao WANG, Feng Ding, Fan Wang, and Hao Li.
\newblock {VTC}-{LFC}: Vision transformer compression with low-frequency
  components.
\newblock In {\em Thirty-Sixth Conference on Neural Information Processing
  Systems}, 2022.

\bibitem{wei2022masked}
Chen Wei, Haoqi Fan, Saining Xie, Chao-Yuan Wu, Alan Yuille, and Christoph
  Feichtenhofer.
\newblock Masked feature prediction for self-supervised visual pre-training.
\newblock In {\em Proceedings of the IEEE/CVF Conference on Computer Vision and
  Pattern Recognition}, pages 14668--14678, 2022.

\bibitem{xie2022simmim}
Zhenda Xie, Zheng Zhang, Yue Cao, Yutong Lin, Jianmin Bao, Zhuliang Yao, Qi
  Dai, and Han Hu.
\newblock Simmim: A simple framework for masked image modeling.
\newblock In {\em Proceedings of the IEEE/CVF Conference on Computer Vision and
  Pattern Recognition}, pages 9653--9663, 2022.

\bibitem{Xu_2020_CVPR}
Kai Xu, Minghai Qin, Fei Sun, Yuhao Wang, Yen-Kuang Chen, and Fengbo Ren.
\newblock Learning in the frequency domain.
\newblock In {\em Proceedings of the IEEE/CVF Conference on Computer Vision and
  Pattern Recognition (CVPR)}, June 2020.

\bibitem{yan2022multiview}
Shen Yan, Xuehan Xiong, Anurag Arnab, Zhichao Lu, Mi Zhang, Chen Sun, and
  Cordelia Schmid.
\newblock Multiview transformers for video recognition.
\newblock In {\em Proceedings of the IEEE/CVF Conference on Computer Vision and
  Pattern Recognition}, pages 3333--3343, 2022.

\bibitem{zeng2020srnet_ECCV}
Ailing Zeng, Xiao Sun, Fuyang Huang, Minhao Liu, Qiang Xu, and Stephen Lin.
\newblock Srnet: Improving generalization in 3d human pose estimation with a
  split-and-recombine approach.
\newblock In {\em ECCV}, 2020.

\bibitem{zhang2022mixste}
Jinlu Zhang, Zhigang Tu, Jianyu Yang, Yujin Chen, and Junsong Yuan.
\newblock Mixste: Seq2seq mixed spatio-temporal encoder for 3d human pose
  estimation in video.
\newblock In {\em Proceedings of the IEEE/CVF Conference on Computer Vision and
  Pattern Recognition}, pages 13232--13242, 2022.

\bibitem{zheng2020deep}
Ce Zheng, Wenhan Wu, Taojiannan Yang, Sijie Zhu, Chen Chen, Ruixu Liu, Ju Shen,
  Nasser Kehtarnavaz, and Mubarak Shah.
\newblock Deep learning-based human pose estimation: A survey, 2020.

\bibitem{Zheng_2021_ICCV}
Ce Zheng, Sijie Zhu, Matias Mendieta, Taojiannan Yang, Chen Chen, and Zhengming
  Ding.
\newblock 3d human pose estimation with spatial and temporal transformers.
\newblock In {\em Proceedings of the IEEE/CVF International Conference on
  Computer Vision (ICCV)}, pages 11656--11665, October 2021.

\bibitem{zhu2020deformable}
Xizhou Zhu, Weijie Su, Lewei Lu, Bin Li, Xiaogang Wang, and Jifeng Dai.
\newblock Deformable detr: Deformable transformers for end-to-end object
  detection.
\newblock {\em arXiv preprint arXiv:2010.04159}, 2020.

\end{thebibliography}
}

\end{document}